\documentclass[jair,twoside,11pt,theapa]{article}
\usepackage{jair, theapa, rawfonts}

\newcommand{\qthd}{Q_\mathcal{D}(y_{1:t-1},  y_t)}
\newcommand{\qthc}{Q_\mathcal{C}(y_{1:t-1}, y_t, c)}
\newcommand{\vthd}{V_\mathcal{D}(y_{1:t-1})}
\newcommand{\vthc}{V_\mathcal{C}(y_{1:t-1,} c)}

\usepackage{amsmath,amssymb}
\usepackage{commath}

\usepackage{url}
\usepackage{amsmath}
\usepackage{amssymb}
\usepackage{amsfonts}

\usepackage[linesnumbered,ruled,vlined]{algorithm2e}
\usepackage[noend]{algpseudocode}
\usepackage{graphicx}
\usepackage{subfig}
\usepackage{caption}
\usepackage{mdwmath}
\usepackage{mdwtab}
\usepackage{booktabs}
\ShortHeadings{Leveraging GPT-2 for Opinion Spam Detection with Limited Labeled Data}
{Irissappane, Yu, Shen, Agrawal \& Stanton}
\firstpageno{1}
\graphicspath{ {Figs} }

\begin{document}

\title{Leveraging GPT-2 for Classifying Spam Reviews with Limited Labeled Data via Adversarial Training\thanks{In Submission at JAIR}}

\author{\name Athirai A. Irissappane \email athirai@uw.edu \\
       \name Hanfei Yu \email hanfeiyu@uw.edu \\
       \name Yankun Shen \email yankun@uw.edu \\
        \name Anubha Agrawal \email anubha04@uw.edu \\
        \addr
        School of Engineering and Technology,\\
        University of Washington, Tacoma
       \AND
       \name Gray Stanton \email gray.stanton@colostate.edu \\
       \addr
       Department of Statistics, \\
       Colorado State University
       }


\maketitle

\begin{abstract}
Online reviews are a vital source of information when purchasing a service or a product. Opinion spammers manipulate these reviews, deliberately altering the overall perception of the service. Though there exists a corpus of online reviews, only a few have been labeled as spam or non-spam, making it difficult to train spam detection models. We propose an adversarial training mechanism leveraging the capabilities of Generative Pre-Training 2 (GPT-2) for classifying opinion spam with limited labeled data and a large set of unlabeled data. Experiments on TripAdvisor and YelpZip datasets show that the proposed model outperforms state-of-the-art techniques by at least $7\%$ in terms of accuracy when labeled data is limited. The proposed model can also generate synthetic spam/non-spam reviews with reasonable perplexity, thereby, providing additional labeled data during training.
\end{abstract}

\section{Introduction} 
\label{Introduction}

Opinion spam is a widespread problem for businesses such as e-commerce, social media, travel sites, and movie review sites~\cite{Jindal2010FindingUR}. Statistics show that over $95\%$ of consumers read reviews before making a purchase\footnote{https://learn.g2crowd.com/customer-reviews-statistics}. The likelihood of a successful purchase is reported to increase when the number of reviews increases. Opinion spammers exploit these trends for financial gains by providing spam reviews, influencing readers, and thereby affecting sales. We will consider the problem of identifying spam reviews as a classification problem, i.e., our goal is to classify a given review as spam or non-spam.

One of the main challenges in identifying spam reviews is the lack of labeled data, i.e., data that is already labeled as spam and non-spam~\cite{Rayana2015}. While there exists a corpus of online reviews, only a few of them are labeled because manual labeling is often time-consuming, costly, and subjective~\shortcite{li2018generative}. There is very limited research on using semi-supervised learning techniques for opinion spam detection~\shortcite{crawford2015survey}, however, results show that unlabeled data, when used in conjunction with small amounts of labeled data can produce a considerable improvement in learning accuracy~\shortcite{Ott2011FindingDO}. Existing semi-supervised learning approaches~\shortcite{li2011learning,hernandez2013using,li2014spotting} to identify opinion spam use a pre-defined set of features to train their classifier. In our work, we will use neural networks (generative adversarial networks), which automatically discover features needed for opinion spam classification~\shortcite{lecun2015deep}. 

Deep generative models, especially Generative Adversarial Networks (GANs) have shown promising results in semi-supervised learning~\shortcite{kumar2017semi}. GANs~\shortcite{goodfellow2014generative} learn to generate samples very close to real data. They operate by training two neural networks that play a min-max game. The discriminator network tries to discriminate real training samples from fake ones and the generator network tries to generate fake training samples to fool the discriminator. However, most research on GANs is for images, i.e., continuous values instead of textual data, which are discrete in nature~\shortcite{fedus2018maskgan}. The main drawbacks in using GANs for text data are: 1) when data is discrete, gradient from discriminator may not be useful for improving generator, because a slight change in weights brought forth by gradients may not correspond to a suitable discrete dictionary mapping~\cite{huszar2015not}; 2) the GAN loss is computed based on the entire sentence and not using a partial one as it is cumbersome to formalize rewards for the entire sentence based on a few initial tokens, leading to the sparse rewards problem~\shortcite{yu2017seqgan}.

To address the above drawbacks, we model the generator as a reinforcement learning agent. The generator learns a policy to select the next token in the sentence based on the long-term rewards, modeling it as a sequential decision-making process. To optimize the generator, we use the actor-critic method~\cite{konda2000actor} where the discriminator provides reward signals to guide the generator. Since our main goal is to classify spam sentences, we also include a classifier component, which additionally provides reward signals to the generator. A Few GAN-based methods such as SeqGAN~\cite{yu2017seqgan}, StepGAN~\cite{tuan2018improving}, and MaskGAN~\cite{fedus2018maskgan}, that model text \textit{generation} as a reinforcement learning process exists. These approaches do not focus on text classification and are also limited by the length of the sentence that can be generated, e.g., MaskGAN can only generate $40$ words per sentence. Therefore, these approaches may not be suitable for online reviews which are relatively lengthy. CSGAN~\cite{li2018generative} focuses on text classification, however, it solely depends on labeled data and uses the time-consuming monte carlo tree search roll-outs to deal with the sparse rewards problem. Further, CSGAN may not be suitable for opinion spam detection due to lengthy reviews, subtlety of classification, computation time, and lack of sufficient labeled data.


In this paper, we propose spamGAN-GPT2, an approach for classifying opinion spam. spamGAN-GPT2 consists of $3$ different components: generator, discriminator, and classifier. These $3$ components work together to not only classify spam reviews but also to generate samples similar to the train set. spamGAN-GPT2 is trained using (limited) labeled and large unlabeled data to accurately capture the input distribution, resulting in better classification accuracy for comparatively longer reviews. Note that spamGAN-GPT2 can be applied to any text classification problem and is not restricted to opinion spam detection. 

While we can use a simple Recurrent Neural Network (RNN) architecture to handle the sequence of tokens in our generator, discriminator and classifier components, performance may degrade due to the vanishing gradients problem, inherent to RNNs. Although Long Short Term Memory (LSTM) and Gated Recurrent Unit (GRU) address this issue, they are still not capable of handling long sequences~\shortcite{RNN-memory-loss}. Attention mechanisms have become hugely popular in sequence modeling as they focus attention on specific parts of the sentence without the need for sequential computation, thereby, capable of handling longer sentences without memory constraints. The transformer architecture~\shortcite{Transformer}, which is solely based on attention mechanism, has outperformed RNNs in learning semantic information, language generation as well as text classification. We will use the transformer architecture for our generator, discriminator, and classifier components. But, instead of training the transformer model from scratch, we will use a pre-trained model.

Pre-trained models have shown to be effective in improving many natural language processing tasks such as text classification, question-answering~\shortcite{devlin2018bert}. Generally, pre-trained models are trained on large amounts of data, using resources that may not be accessible to everyone. Multiple transformer-based pre-trained models have been proposed in recent years. One such pre-trained model is Generative Pre-Training 2 (GPT-2), which is a large Transformer-based language model with $117$ million parameters, trained on a dataset of $8$ million web pages~\shortcite{gpt-2}. GPT-2 has achieved state-of-the-art results on a variety of language modeling tasks~\shortcite{gpt-2}. We will use the pre-trained GPT-2 model in our approach by fine-tuning it to our opinion-spam dataset as it will converge much faster and achieve better performance than training from scratch.


The proposed approach spamGAN-GPT2\footnote{Source code can be found at https://github.com/airesearchuwt/spamGAN} is based on our previous spamGAN work~\shortcite{spamGAN}. We make the following additional contributions in this paper: 1) we extend the previous spamGAN approach using the pre-trained GPT-2 model for better classification accuracy as well as generated sentence quality. To the best of our knowledge, we are the first to explore GANs with a pre-trained transformer-based model for opinion spam detection; 2) in the previous spamGAN model, all three components, i.e., the generator, discriminator, and classifier are implemented using RNNs, which can only handle input sentences up to a length of $128$ tokens. The proposed spamGAN-GPT2 model can process sentences over $400$ tokens; 3) we propose a customized stochastic teacher-forcing decoding strategy, which improves the robustness of the reinforcement learning process during adversarial training; 4) we conduct additional experiments on the YelpZip dataset. Extensive evaluation on both TripAdvisor and YelpZip datasets demonstrate that the proposed spamGAN-GPT2 outperforms the state-of-the-art opinion spam detection methods by at least $7\%$\footnote{The values are lower bound on the improvements obtained on TripAdvisor and YelpZip datasets.} in terms of accuracy and the previous spamGAN approach by $5.86\%$. The perplexity of the generated sentences improves by $82.4\%$ when compared to spamGAN.

The rest of the paper is organized as follows. Sec.~\ref{sec:relatedwork} describes the related research on opinion spam detection. Sec.~\ref{sec:methodology} describes the previous spamGAN approach and the proposed spamGAN-GPT2 model in detail. Sec.~\ref{sec:experiments} shows our evaluation results on TripAdvisor and YelpZip datasets. Finally, Sec.~\ref{sec:conclusion} concludes the paper with directions for future work.

\section{Related Work}
\label{sec:relatedwork}

\subsection{Opinion Spam Detection}
Existing opinion spam detection techniques are mostly supervised with pre-defined features. Jindal et al.~\citeyear{Jindal2008OpinionSA} used logistic regression with product, review, and reviewer-centric features. Ott et al.~\citeyear{Ott2011FindingDO} used n-gram features to train a Naive Bayes and SVM classifier. Feng et al.~\citeyear{Feng2012SyntacticSF}, Mukerjee et al.~\citeyear{mukherjee2013yelp}, Li et al.~\citeyear{li2015analyzing} used part-of-speech tags with context-free grammar parse trees, behavioral, and spatio-temporal features, respectively.

Deep learning techniques for spam detection directly consider the reviews as input without a separate feature extraction process. GRNN~\shortcite{ren2017neural} used a gated recurrent neural network to study the contextual information of reviews. DRI-RCNN~\shortcite{zhang2018dri} used a recurrent network for learning the contextual information of the words in the reviews. DRI-RCNN extends RCNN~\shortcite{lai2015recurrent} by learning embedding vectors with respect to both spam and non-spam labels. As RCNN and DRI-RCNN use neural networks for classification, we will compare with these supervised methods in our experiments.

Few semi-supervised methods for opinion spam detection exist. Li et al.~\citeyear{li2011learning} used co-training with Naive-Bayes classifier on reviewer, product, and review features. Hern\'andez et al.~\citeyear{hernandez2013using} and Li et al.~\citeyear{li2014spotting} used only positively labeled samples along with unlabeled data for training. Rayana et al.~\citeyear{Rayana2015} used review features, timestamp, ratings as well as pairwise Markov random field network of reviewers and products to develop a supervised algorithm along with semi-supervised extensions. Some un-supervised methods for spam detection~\shortcite{xu2015unified} exist, but they are out of the scope of this work.

\subsection{GANs for Text Classification}
With respect to GANs for text classification, SeqGAN~\shortcite{yu2017seqgan} addresses the problem of sparse rewards by considering sequence generation as a Reinforcement Learning (RL) problem. Monte Carlo Tree Search (MCTS) is used to overcome the issue of sparse rewards, however, MCTS rollouts are computationally intractable. StepGAN~\shortcite{tuan2018improving} and MaskGAN~\shortcite{fedus2018maskgan} use the actor-critic~\shortcite{konda2000actor} method to learn the rewards, but they are limited by the length of the sequence. Further, all of them focus on text generation. CSGAN~\shortcite{li2018generative} solves text classification problems and incorporates a classifier in its architecture, but, its performance significantly degrades with sentence length as it uses MCTS and character-level embeddings. Our proposed approach differs from CSGAN in using the actor-critic reinforcement learning method for sequence generation and word-level embeddings, suitable for longer sentences. To explain, CSGAN uses MCTS rollouts to deal with the sparse rewards problem, i.e., the rollouts are used to generate the remaining tokens (of the partially generated sentence) as gradient updates are performed based on the score received for the entire sentence. Though CSGAN uses character-level embeddings, which reduce the action space from which the next token is sampled, it is still a time-consuming process. Our proposed method, on the other hand, does not use MCTS rollouts and estimates the rewards requiring only a single pass of the generated sentence through the discriminator and classifier, thereby reducing training time. 

\subsection{Pre-Training in NLP}


Unlike machine learning techniques which directly learn each task from the scratch, transfer learning techniques transfer knowledge from previously learned tasks to a target task when the latter has fewer high-quality training samples~\shortcite{Pan10asurvey}. Transfer learning involves: 1) pre-training a model on a given task. In the case of Natural Language Processing (NLP), it is common to pre-train word embeddings, which capture syntactic and semantic information of words from a large corpus of texts~\shortcite{turian-etal-2010-word} and 2) adapting the learned model to a new but similar task via fine-tuning~\shortcite{tune2019}, in which the pre-trained model is updated to suit the new task~\shortcite{devlin2018bert}. In NLP, fine-tuned pre-trained word-embeddings outperform traditional word embedding methods that need a large corpus of data for training from scratch. Further, the pre-trained models need to completely re-learn only a few parameters while retaining or slightly modifying the others. 

In 2017, the transformer architecture~\shortcite{Transformer} was introduced for language translation solely based on attention mechanisms without recurrence and convolution layers. The transformer consists of encoder and decoder components, each including self-attention modules, resulting in a highly parallelizable architecture capable of handling longer sentences. The transformer model has shown better performance in language understanding, learning semantic information as well as language generation tasks when compared to traditional RNNs. Multiple pre-trained transformer-based models have been proposed in recent years. Some of them such as Bidirectional Encoder Representations from Transformers (BERT)~\cite{devlin2018bert} are completely based on transformer encoder, while others like Generative Pre-Training (GPT-2)~\shortcite{gpt-2} use the transformer decoder. 

Auto-regressive language models are the ones in which every token generated (as a part of the sentence) has the context of the previous tokens. In general, auto-regressive models are only trained to encode uni-directional context, i.e., either forward or backward. In contrast, the auto-encoding language models like BERT use the entire surrounding context at once. BERT relies on corrupting the input with masks~\shortcite{yang2019xlnet} and has the ability to process bidirectional contexts. However, BERT neglects the dependency between the masked positions and suffers from a pre-train fine-tune discrepancy~\shortcite{yang2019xlnet}. The GPT-2 model, on the other hand, is auto-regressive in nature. GPT-2 is pre-trained on the standard task of predicting the next token, given a sequence of prior tokens. It has also achieved state-of-the-art results on $7$ out of $8$ tested language modeling datasets~\shortcite{gpt-2}. As our proposed text classification model is auto-regressive in nature and encompasses the word embedding learning task which is a natural fit for GPT-2, we will use the pre-trained GPT-2 model (instead of BERT) in our work. 

GAN-BERT~\shortcite{gan-bert} was recently introduced to extend the BERT model for semi-supervised learning tasks. They mainly use BERT to improve the classification performance by learning better sentence embeddings from both the labeled and unlabeled data. However, due to the limited generation ability of BERT, the generator of GAN-BERT benefits little from transformer architectures. Unlike GAN-BERT, the generator of our proposed spamGAN-GPT2 is auto-regressive nature and the discriminator and classifier are auto-encoding models. These components work towards improving each other, i.e., we see a significant improvement in the text classification as well as the quality of the synthetic reviews produced by the generator (see experimental results in Sec.~\ref{sec:experiments} for more details).

\section{Proposed Approach}\label{sec:methodology}
We will first introduce the spamGAN approach, which uses recurrent layers for processing the sequence of tokens in Sec.~\ref{sec:spamGANsec}. We will then describe how GPT-2 is used to replace the recurrent layers in Sec.~\ref{sec:spamgpt2sec}.

\subsection{spamGAN}\label{sec:spamGANsec}


Let $\mathbb{D_L}$ be the labeled set of reviews, labeled either spam or non-spam. Given the cost of labeling, we hope to improve the classification performance by also using $\mathbb{D_U}$, which is a significantly larger set of unlabeled reviews\footnote{$\mathbb{D_U}$ includes both spam/non-spam reviews.}. Let $\mathbb{D}=\mathbb{D_L} \cup \mathbb{D_U}$ be a combination of both labeled and unlabeled sentences for training\footnote{Only $\mathbb{D_L}$ or both $\mathbb{D_L}$ and $\mathbb{D_U}$ can be used during training (see Alg.~\ref{alg:spamgan}).}. Each sentence $y_{1:T}= \{y_1, y_2, \ldots y_t, \ldots, y_T\}$ in the training set consists of a sequence of $T$ word tokens, where $y_t \in \mathtt{Y}$ represents the $t^{th}$ token in the sentence and $\mathtt{Y}$ is a corpus of all tokens used. For sentences belonging to $\mathbb{D_L}$, we also include a class label from one of the two classes\footnote{We will use $\mathfrak{c}$ to represent the true class label and $c$ to represent the sampled/predicted class label.} $ \mathfrak{c} \in \mathbb{C}:\{\mathtt{spam}, \mathtt{non\text{-}spam}\}$.

\begin{figure}[h]
\centering
  \includegraphics[height=4cm, keepaspectratio]{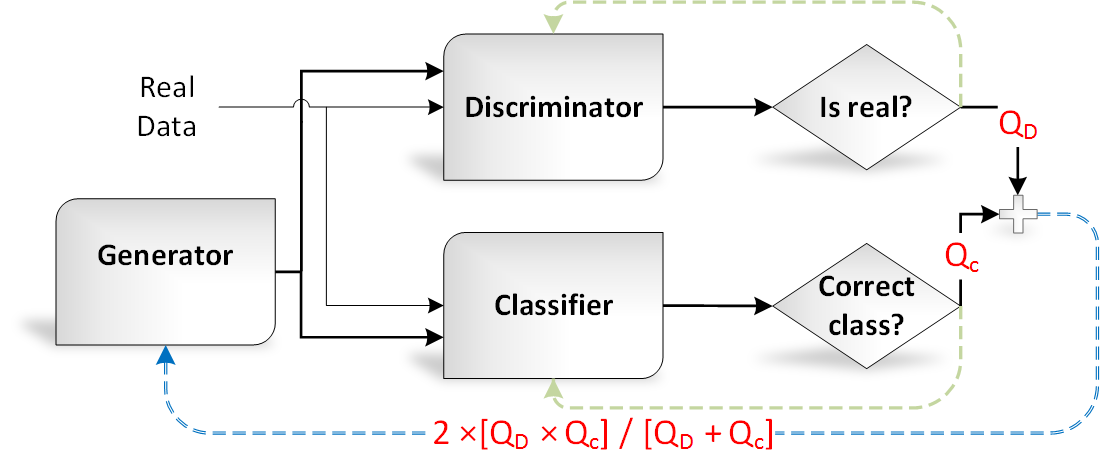}
  \caption{spamGAN Architecture}
  \label{fig:architecture}
\end{figure}

SpamGAN leverages both the labeled and unlabeled data and consists of three components: the generator $\mathcal{G}$, the discriminator $\mathcal{D}$, and the classifier $\mathcal{C}$ as shown in Fig.~\ref{fig:architecture}. The generator, for a given class label, learns to generate new sentences (we call them $\mathtt{fake}$ sentences\footnote{Fake sentences are those produced by the generator. Spam sentences are deceptive sentences with class label $\mathtt{spam}$. Generator can generate fake sentences belonging to $\{\mathtt{spam}$ or $\mathtt{non\text{-}spam}\}$ class.}). These fake sentences are similar to the real ones present in the training set. The discriminator learns to differentiate between real and fake sentences and informs the generator (via rewards) whether the generated sentences are unrealistic. This competition between the generator and discriminator improves the quality of the generated sentence. 

We know the class labels for the fake sentences produced by the generator as they are controlled~\shortcite{hu2017toward}, i.e., constrained by the class labels $\{\mathtt{spam}, \mathtt{non\text{-}spam}\}$. The classifier is trained using both the real labeled sentences from $\mathbb{D_L}$ and fake sentences produced by the generator. This improves the ability of the classifier to generalize beyond the small set of labeled sentences. The classifier's performance on fake sentences from the generator is also used as feedback to improve the generator; better classification accuracy results in more rewards. While the discriminator and generator are competing, the classifier and generator are mutually bootstrapping. As the three components of spamGAN are trained, the generator produces sentences very similar to the training set while the classifier learns the characteristics of spam and non-spam sentences to differentiate them correctly. 

\begin{figure}[h]
\centering
  \includegraphics[height=5.5cm, keepaspectratio]{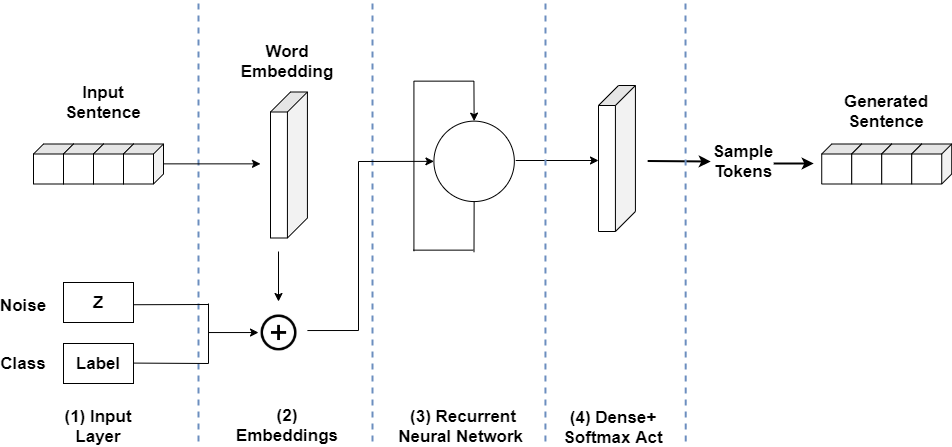}
    \caption{SpamGAN Generator Architecture}
    \label{fig:spamgangen}
\end{figure}

\vspace{2mm}
\noindent 
{\bf{Generator.}}
If $P_R(y_{1:T}, \mathfrak{c})$ is the true joint distribution of sentences $y_{1:T}$ and classes $\mathfrak{c} \in \mathbb{C}$ from the real training set, the generator aims to find a parameterized conditional distribution $\mathcal{G}(y_{1:T} | z, c,\theta_g)$ that best approximates the true distribution. The generated fake sentence is conditioned on the network parameters $\theta_g$, the noise vector $z$, and the class label $c$, which are sampled from the priors $P_z$ and $P_{\mathfrak{c}}$, respectively. The context vector, consisting of $z$, $c$, is concatenated to the (partial) input sentence at every timestep~\cite{tuan2018improving}. This ensures that the actual class label for every generated fake sentence is retained.

 While sampling from $\mathcal{G}(y_{1:T} | z, c,\theta_g)$, the word tokens are generated auto-regressively, decomposing the distribution over token sequences into the ordered conditional sequence.

\vspace{-2mm}
\begin{equation}
\vspace{-0.5mm}
\mathcal{G}(y_{1:T} | z, c, \theta_g) = \prod_{t=1}^{T} \mathcal{G}(y_t| y_{1:t-1}, z, c, \theta_g)
\label{eqn:gen-sample}
\end{equation}

During pre-training, we use batches of real sentences from $\mathbb{D}$ and minimize the cross-entropy of the next token conditioned on the preceding ones. Specifically, we minimize the loss (Eqn.~\ref{eqn:genmlegrad}) over real sentence-class pairs $(y_{1:T}, \mathfrak{c})$ from $\mathbb{D_L}$ as well as unlabeled real sentences from $\mathbb{D_U}$ with randomly assigned class labels drawn from the class prior distribution.

\vspace{-2.5mm}
\begin{equation}
\vspace{-0.5mm}
\mathcal{L}^{\mathcal{G}}_{MLE} = - \sum_{t=1}^{T} \log{\mathcal{G}(y_t | y_{1:t-1}, z, c, \theta_g)} \label{eqn:genmlegrad}
\end{equation}

During adversarial training, we treat sequence generation as a sequential decision-making problem~\cite{yu2017seqgan}. The generator acts as a reinforcement learning agent, trained to maximize the expected rewards using policy gradients. Rewards are the feedback obtained from the discriminator and the classifier for the generated sentences. For implementing the generator, we use a unidirectional multi-layer Recurrent Neural Network (RNN) with gated recurrent units as the base cell as shown in Fig.~\ref{fig:spamgangen}.
  
\begin{figure}[h]
\centering
  \includegraphics[height=5.5cm, keepaspectratio]{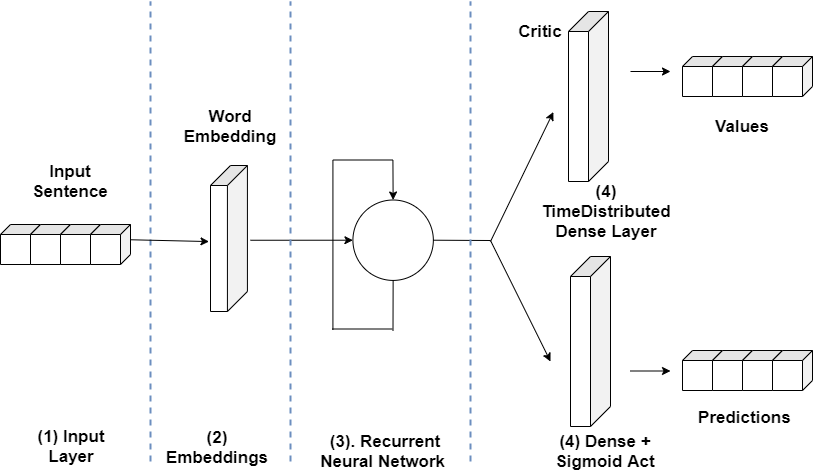}
    \caption{SpamGAN Discriminator/Classifier Architecture}%
    \label{fig:spamgandes}%
\end{figure}

\vspace{2mm}
\noindent 
{\bf{Discriminator.}}
The discriminator $\mathcal{D}$, with parameters $\theta_d$ predicts if a sentence is real (sampled from $P_R$) or fake (produced by the generator) by computing a probability score $\mathcal{D}(y_{1:T}|\theta_d)$ that indicates realness of the sentence. Like~\cite{tuan2018improving}, instead of computing the score at the end of the sentence, the discriminator produces scores $\qthd$ for every timestep, which is then averaged to produce an overall score for the sentence. 

\vspace{-2.5mm}
\begin{equation}
\vspace{-2.0mm}
\mathcal{D}(y_{1:T}|\theta_d) = \frac{1}{T} \sum_{t=1}^T  \qthd 
\label{eqn:dqvalue}
\end{equation}
where, $\qthd$ is the intermediate score for timestep $t$ and is based solely on the preceding partial sentence $y_{1:t}$. In a setup reminiscent of $Q$-learning, we consider $\qthd$ to be the estimated value for the next generated token (action) $y_t$, given the previous tokens (state) $y_{1:t-1}$. 
Thus, the discriminator provides estimates for the true state-action values without the additional computational overhead of using MCTS rollouts.  

We train the discriminator by maximizing the score $\mathcal{D}(y_{1:T}|\theta_d)$ for the real sentences and minimizing it for fake ones. This is achieved by minimizing the loss $\mathcal{L^{(D)}}$ given below.

\begin{equation}
\begin{split}
\hspace{-2.5mm} \mathcal{L^{(D)}} \hspace{-1.5mm} = \hspace{-2.5mm} & \hspace{-1.5mm} \mathop{\mathbb{E}}_{y_{1:T} \sim P_R} \hspace{-3.5mm}  -\sbr{\log \mathcal{D}(y_{1:T}|\theta_d)} \hspace{-0.5mm} + \hspace{-1.5mm} \mathop{\mathbb{E}}_{y_{1:T} \sim \mathcal{G}}\hspace{-1.5mm} - \hspace{-0.5mm} \sbr{\log{(1 \hspace{-1.5mm}- \hspace{-1.5mm}\mathcal{D}(y_{1:T}|\theta_d))}} 
\label{eqn:dloss}
\end{split}
\end{equation}

We also include a discrimination critic $\mathcal{D}_{crit}$~\cite{konda2000actor}, which is trained to approximate the score $\qthd$ from the discriminator network, for the preceding partial sentence $y_{1:t-1}$. The approximated score $\vthd$ will be used as a value baseline to stabilize the policy gradient updates for the generator during adversarial training. 

\begin{equation}
 \vthd = \mathop{\mathbb{E}}_{y_t} \sbr{\qthd}
\end{equation}

The discrimination critic $\mathcal{D}_{crit}$ is trained to minimize the sequence mean-squared error between $\vthd$ and the actual score $\qthd$ as shown in the equation below,

\vspace{-1mm}
\begin{equation}
\vspace{-1mm}
\begin{split}
\mathcal{L^{(D_\text{crit})}} & = \mathop{\mathbb{E}}_{y_{1:T}} \sum_{t=1}^{T} \norm{\qthd - \vthd}^2 
\label{eqn:dcriticloss}
\end{split}
\end{equation}

 The discriminator network is implemented as a unidirectional RNN followed by a dense layer with sigmoid activation as shown in Fig.~\ref{fig:spamgandes}. It generates the probability that a sentence is real at every timestep, i.e., $\qthd$. For the discrimination critic, we have an additional head (different from the one that computes $\qthd$) attached to the discriminator RNN, which estimates $V_{\mathcal{D}}(y_{1:t-1})$ for each timestep.

\vspace{2mm}
\noindent 
{\bf{Classifier.}}
Given a sentence $y_{1:T}$, the classifier $\mathcal{C}$ with parameters $\theta_c$ predicts if the sentence belongs to class $c \in \mathbb{C}$. Like the discriminator, the classifier assigns a prediction score at every timestep $\qthc$ for the partial sentence $y_{1:t}$, which identifies the probability of the sentence belonging to class $c$. The intermediate scores at every timestep are then averaged to produce the overall score given by,
\begin{equation}
\begin{split}
\mathcal{C}(y_{1:T}, c|\theta_c) & = \frac{1}{T} \sum_{t=1}^T \qthc  
\label{eqn:cqvalue}
\end{split}
\end{equation}

The classifier loss $\mathcal{L^{C}}$ is based on: 1) $\mathcal{L^{(C_{\text{R}})}}$, the cross-entropy loss for real labeled sentences computed using the overall classifier sentence score; 2) $\mathcal{L^{(C_{\text{G}})}}$, the loss for the fake sentences. Fake sentences are considered as potentially noisy training examples, so we not only minimize cross-entropy loss but also include Shannon entropy $\mathcal{H}(\mathcal{C}(c|y_{1:T},\theta_C))$. 


\vspace{-1mm}
\begin{equation}
\begin{split}
\mathcal{L^{C}} & = \mathcal{L^{(C_{\text{R}})}} + \mathcal{L^{(C_{\text{G}})}}\\ \label{eqn:closs}  \\
\mathcal{L^{(C_{\text{R}})}}& =  \mathop{\mathbb{E}}_{(y_{1:T}, c) \sim P_R(y, \mathfrak{c})}\sbr{-\log \mathcal{C}(c|y_{1:T},\theta_c)} \\ 
\mathcal{L^{(C_{\text{G}})}}  & =  \mathop{\mathbb{E}}_{ c \sim P_c, y_{1:T} \sim \mathcal{G}} [-\log \mathcal{C}(c|y_{1:T},\theta_c) - \beta \mathcal{H}(\mathcal{C}(c|y_{1:T},\theta_C))]
\end{split}
\end{equation}

In $\mathcal{L^{(C_{\text{G}})}}$, $\beta$, the balancing parameter, influences the impact of Shannon entropy. Including $\mathcal{H}(\mathcal{C}(c|y_{1:T},\theta_C))$, for minimum entropy regularization~\cite{hu2017toward}, allows the classifier to predict classes for generated fake sentences more confidently. This is crucial in reinforcing the generator to produce sentences of the given class during adversarial training. 

Like the discriminator, we include a classification critic $\mathcal{C}_{crit}$ to estimate the classifier score $\qthc$ for the preceding partial sentence $y_{1:t-1}$, 

\begin{equation}
\vthc = \mathop{\mathbb{E}}_{y_t}[\qthc]
\end{equation}

The implementation of the classifier is similar to the discriminator as shown in Fig.~\ref{fig:spamgandes}. We use a unidirectional RNN with a dense output layer producing the predicted probability distribution over classes $c \in \mathbb{C}$. The classification critic is also an alternative head off the classifier RNN with an additional dense layer estimating $\vthc$ for each timestep. We train this classification critic by minimizing $\mathcal{L^{(C\text{crit})}}$,

\vspace{-1mm}
\begin{equation}
\begin{split}
\mathcal{L^{(C_\text{crit})}} & = \mathop{\mathbb{E}}_{y_{1:T}} \sum_{t=1}^{T} \norm{\qthc - \vthc}^2  
\label{eqn:ccriticloss}
\end{split}
\end{equation}


\vspace{2mm}
\noindent 
{\bf{Reinforcement Learning Component.}}\label{sec:rl}
To address the drawback of sparse rewards, we consider a sequential decision-making framework in which the generator acts as a Reinforcement Learning (RL) agent. In RL, an agent tries to learn the optimal policy in order to take the best possible action for a given state based on the expected rewards (called value). In this problem, the current state of the agent is the tokens generated so far, $s_t = y_{1:t-1}$. The action $y_{t}$ is the next token to be generated, which is selected based on the stochastic policy $\mathcal{G}(y_t| y_{1:t-1}, z, c, \theta_g)$.
The reward the agent receives for the generated sentence $y_{1:T}$ of a given class $c$ is determined by the discriminator and classifier. Specifically, we take the overall scores $\mathcal{D}(y_{1:T}| \theta_d)$ (Eqn.~\ref{eqn:dqvalue}) and $\mathcal{C}(y_{1:T}, c| \theta_c)$ (Eqn.~\ref{eqn:cqvalue}) and blend them in a manner reminiscent of the F1 score~\shortcite{li2018generative}, producing the sentence reward,

\begin{equation}
R(y_{1:T}) = 2\cdot \frac{\mathcal{D}(y_{1:T} | \theta_d) \cdot \mathcal{C}(y_{1:T}, c | \theta_c)}{\mathcal{D}(y_{1:T} | \theta_d) + \mathcal{C}(y_{1:T}, c | \theta_c)} \label{blendedreturn}
\end{equation}

This reward $R(y_{1:T})$ is for the entire sentence delivered during the final timestep, with the reward for every other timestep being zero. Thus, the generator agent seeks to maximize the expected reward, given by, 

\begin{equation}
\begin{split}
\mathcal{L^{(G)}} & = \mathop{\mathbb{E}}_{y_{1:T} {\sim} \mathcal{G}}\sbr{R(y_{1:T})}
\end{split}
\label{eqn:expectreward}
\end{equation}

To maximize $\mathcal{L^{(G)}}$, the generator parameters $\theta_g$ are updated via policy gradients~\shortcite{sutton2000policy}. Specifically, we use the advantage actor-critic method to solve for the optimal policy~\shortcite{konda2000actor}. The expectation in Eqn.~\ref{eqn:expectreward} can be re-written using rewards for intermediate timesteps from the discriminator and classifier. The intermediate scores $\qthd$ from discriminator and $\qthc$ from the classifier are combined as shown in Eqn.~\ref{eqn:rewardspertimestep} 
and serve as estimators for $Q(y_{1:t}, c)$, i.e., the expected reward for $y_{1:t}$.

\begin{equation}
\begin{split}
& Q(y_{1:t}, c)  = 2 \cdot \frac{\qthd \cdot \qthc}
{\qthd +\qthc}\\
& V(y_{1:t-1},c) = 2 \cdot \frac{\vthd \cdot  \vthc}
{\vthd + \vthc} 
\label{eqn:rewardspertimestep} 
\end{split}
\end{equation}

During adversarial training, we perform gradient ascent to update the generator using Eqn.~\ref{eqn:deltag}. To reduce variance in the gradient estimates, we replace $Q(y_{1:t}, c)$ by the advantage function $Q(y_{1:t}, c) - V(y_{1:t-1}, c)$, where $V(y_{1:t-1}, c)$ is given in Eqn.~\ref{eqn:rewardspertimestep}. We use $\alpha = T - t$ in Eqn.~\ref{eqn:deltag} to increase the importance of initially generated tokens while updating $\theta_g$. $\alpha$ is a linearly decreasing factor that corrects the relative lack of confidence in the initial intermediate scores from discriminator and classifier. 

\vspace{-2.5mm}
\begin{align}
\nabla_{\theta_g} \mathcal{L^{(G)}}  = \mathop{\mathbb{E}}_{y_{1:T}} \sum_t^T & \alpha \sbr{Q(y_{1:t}, c) - V(y_{1:t-1}, c)}  \times \nabla_{\theta_g} \log \mathcal{G}(y_t | y_{1:t-1}, z, c, \theta_g) 
  \label{eqn:deltag}
  \vspace{-3.5mm}
\end{align}
\vspace{-5.5mm}

\begin{algorithm}[t]
\small{
\caption{spamGAN}
\label{alg:spamgan}
\textbf{Input}: Labeled dataset $\mathbb{D_L}$, Unlabeled dataset $\mathbb{D_U}$ \\
\textbf{Parameters}: Network parameters $\theta_g\; \theta_d \; \theta_c \; \theta_{dcrit} \; \theta_{ccrit}$ \\
Perform pre-training\\
\For{$\mathtt{Training\text{-}epochs}$}{
\For{$\mathtt{G\text{-}Adv\text{-}epochs}$}{
sample batch of classes $c$ from $\sim P_{\mathfrak{c}}$ \\
generate batch of fake sentences $y_{1:T} \sim \mathcal{G}$ given $c$\\
\For{$t \in 1:T$}{
compute $Q(y_{1:t}, c)$, $V(y_{1:t-1},c)$ using Eqn.~\ref{eqn:rewardspertimestep}
}
update $\theta_g$ using policy gradient $\nabla_{\theta_g} \mathcal{L^{(G)}}$ in Eqn.~\ref{eqn:deltag}\\
}
\For{$\mathtt{G\text{-}MLE\text{-}epochs}$}{
sample batch of real sentences from $\mathbb{D_L}$, $\mathbb{D_U}$ \\
Update $\theta_g$ using MLE in Eqn.~\ref{eqn:genmlegrad} \\
}

\For{$\mathtt{D\text{-}epochs}$}{
sample batch of real sentences from $\mathbb{D_L}$, $\mathbb{D_U}$ \\
sample batch of fake sentences from $\mathcal{G}$  \\
update discriminator using $\nabla_{\theta_d} \mathcal{L^{(D)}}$ from Eqn.~\ref{eqn:dloss}\\
compute $\qthd, \vthd$ for fake sentences \\
update $\mathcal{D}_{\text{crit}}$ using $\nabla_{\theta_{dcrit}} \mathcal{L^{(D\text{crit})}}$ from Eqn.~\ref{eqn:dcriticloss} \\
}
\For{$\mathtt{C\text{-}epochs}$}{
sample batch of real sentences-class pairs from $\mathbb{D_L}$ \\
sample batch of fake sentence-class pairs from $\mathcal{G}$ \\
update classifier using $\nabla_{\theta_c} \mathcal{L^{(C)}}$ from Eqn.~\ref{eqn:closs} \\
\hspace{-2mm} compute \hspace{-1mm} $\qthc,\hspace{-0.5mm}\vthc$ on fake sentences \\
update $\mathcal{C}_{\text{crit}}$ using $\nabla_{\theta_{ccrit}} \mathcal{L^{(C\text{crit})}}$ from Eqn.~\ref{eqn:ccriticloss}
}
}
}
\vspace{-1.5mm}
\end{algorithm}

\vspace{2mm}
\noindent 
{\bf{spamGAN algorithm.}}\label{sec:pre}
Alg.~\ref{alg:spamgan} describes the spamGAN approach in detail. Before beginning adversarial training, we pre-train the different components of spamGAN. The generator $\mathcal{G}$ is pre-trained using maximum likelihood estimation (MLE)~\shortcite{grover2018flow} by updating the parameters via Eqn.~\ref{eqn:genmlegrad}. Once the generator is pre-trained, we take batches of real sentences from the labeled dataset $\mathbb{D_L}$, the unlabeled dataset $\mathbb{D_U}$ and fake sentences sampled from the generator $\mathcal{G}(y_{1:T} | z, c, \theta_g)$ to pre-train the discriminator minimizing the loss $\mathcal{L^{(D)}}$ in Eqn.~\ref{eqn:dloss}. The classifier $\mathcal{C}$ is pre-trained solely on real sentences from the labeled dataset $\mathbb{D_L}$. It is trained to minimize the cross-entropy loss $\mathcal{L^{(C_{\text{R}})}}$ on real sentences and their labels (Eqn.~\ref{eqn:closs}). The discrimination and classification critics $\mathcal{D_{\text{crit}}}$ and $\mathcal{C_{\text{crit}}}$ are trained by minimizing their loses $\mathcal{L^{(D\text{crit})}}$ (Eqn.~\ref{eqn:dcriticloss}) and $\mathcal{L^{(C\text{crit})}}$ (Eqn.~\ref{eqn:ccriticloss}), respectively. Such pre-training addresses the problem of mode collapse~\shortcite{guo2018long} to a satisfactory extent.

 After pre-training, we perform the adversarial training for $\mathtt{Training\text{-}epochs}$ (Lines $4$-$25$). We create a batch of fake sentences using the generator $\mathcal{G}$ by sampling the classes $c$ from a prior distribution $P_{\mathfrak{c}}$ (Lines $6$-$7$). We compute $Q(y_{1:t}, c)$, $V(y_{1:t-1},c)$ using Eqn.~\ref{eqn:rewardspertimestep} for every timestep (Line $9$). The generator is then updated using the policy gradient in Eqn.~\ref{eqn:deltag} (Line $10$). This process is repeated for $\mathtt{G\text{-}Adv\text{-}epochs}$. Like~\shortcite{Li2017}, the training robustness is greatly improved when the generator is updated using MLE through Eqn.~\ref{eqn:genmlegrad} on sentences from $\mathbb{D}$ (Lines $11$-$13$). We then train the discriminator by sampling real sentences from $\mathbb{D_L}$, $\mathbb{D_U}$ as well as the fake sentences from the generator (Lines $15$-$16$). The discriminator is updated using Eqn.~\ref{eqn:dloss} (Line $17$). We also train the discrimination critic, by computing $\qthd, \vthd$ for the fake sentences and updating the gradients using Eqn.~\ref{eqn:dcriticloss} (Line $18$-$19$). This process is repeated for $\mathtt{D\text{-}epochs}$. We perform a similar set of operations for the classifier for $\mathtt{C\text{-}epochs}$ (Lines $20$-$25$). 


\subsection{spamGAN-GPT2}\label{sec:spamgpt2sec}

We extend the above spamGAN approach using the pre-trained GPT-2 model~\shortcite{gpt-2} for better classification accuracy as well as generated sentence quality.
The proposed spamGAN-GPT2 also includes three components, generator $\mathcal{G}$, discriminator $\mathcal{D}$, and classifier $\mathcal{C}$. However, we replace the RNN layer using GPT-2, as explained below.  


\begin{figure}[h]
    \centering
    \subfloat[GPT-2 Transformer Decoders]{\includegraphics[height=4.5cm, keepaspectratio]{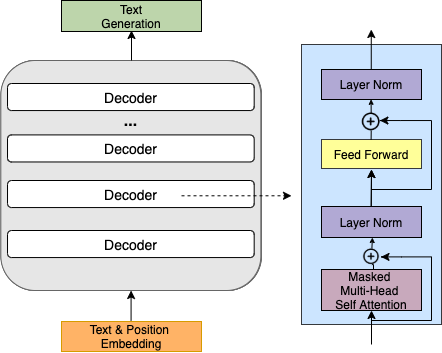}
\label{fig:gpt2decoder}}%
    \qquad
    \subfloat[Adapted GPT-2 Transformer Encoders]{\includegraphics[height=4.5cm, keepaspectratio]{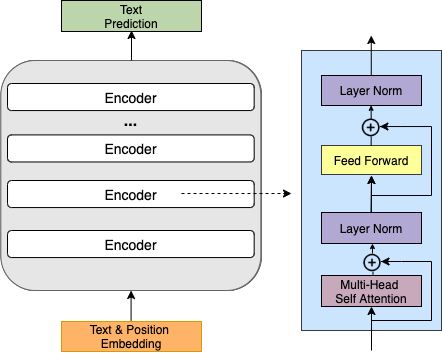}
\label{fig:gpt2encoder}}%
    \caption{GPT-2 Architecture \shortcite{gpt}}%
    \label{fig:gpt2}%
\end{figure}





\vspace{2mm}
\noindent 
{\bf{Generative Pre-Trained-2 (GPT-2).}}
GPT-2 is a pre-trained, transformer-based language model that is trained on a massive dataset containing 8-million web-pages. GPT-2 is a direct scale-up of GPT~\shortcite{gpt}, with more than 10 times the parameters and trained on more than 10 times the amount of data~\shortcite{gpt-2}. It achieves state-of-the-art performance on many benchmark language modeling tasks. It can perform reading comprehension, machine translation, question answering, language generation as well as summarization tasks~\shortcite{gpt-2}. We will use the smaller version of GPT-2, which consists of $12$ transformer decoders as shown in Fig.~\ref{fig:gpt2decoder} due to resource constraints. This smaller GPT-2 version consists of $117M$ parameters with $768$ dimension word embedding and $1,024$ dimension position embedding. The $12$ transformer decoder blocks consist of $12$ multi-head masked self-attention layers with $768$ units and $12$ feed-forward layers with $3072$ units. The vocabulary size used is $50,257$. Instead of RNNs, in the generator component of the proposed spamGAN-GPT2, we will use the GPT-2 transformer decoders as shown in Fig.~\ref{fig:gpt2decoder}. For the discriminator and classifier components, we will replace RNNs with a similar GPT-2 architecture, however, we will use a stack of $12$ transformer encoders instead of decoders with multi-head attention as shown in Fig.~\ref{fig:gpt2encoder}.

\begin{figure}[t]
\centering
 \includegraphics[height=6.5cm, keepaspectratio]{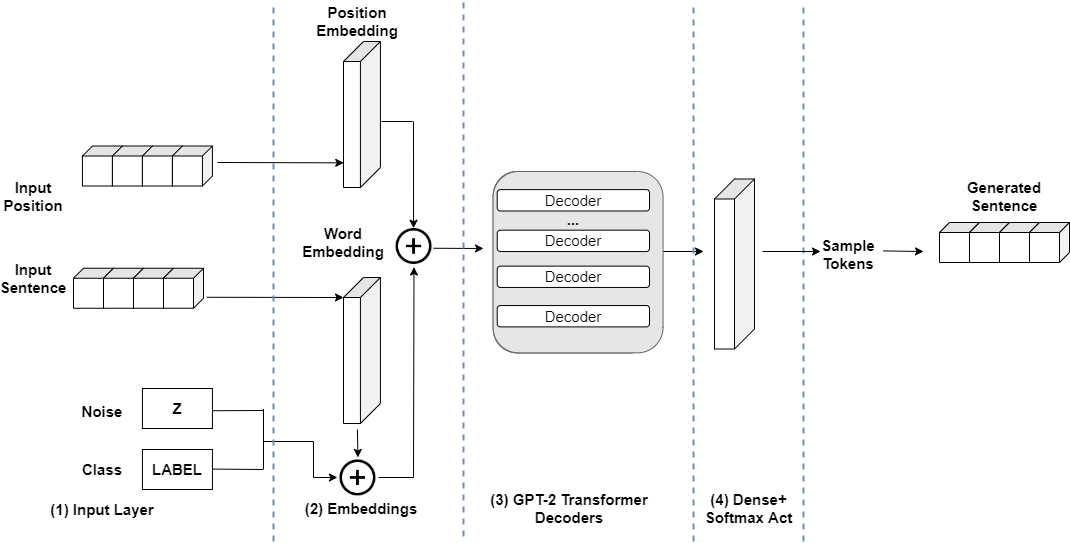}
\caption{spamGAN-GPT2 Generator Architecture}
\label{fig:spamgangpt2-generator}
\end{figure}

\vspace{2mm}
\noindent 
{\bf{Generator.}}
The generator component of spamGAN-GPT2 follows the same architecture as spamGAN. However, it replaces the RNN layer (in Fig.~\ref{fig:spamgangen}) with GPT-2 as shown in Fig.~\ref{fig:spamgangpt2-generator}. Here, the generator additionally takes as input the position information $\mathfrak{p}$ of the tokens in the sentence to produce an output distribution $\mathcal{G}(y_t| y_{1:t-1}, \mathfrak{p}, z, c, \theta_g)$ over the target tokens. Here, $y_{1:t-1}$ represents a sequence of $t-1$ tokens, $z$ is the noise vector; $c$ is the class label, and $\theta_g$ represents the parameters of the generator network. The auto-regressive sampling process for the generator is shown in the equation below,

\vspace{-2mm}
\begin{equation}
\vspace{-0.5mm}
\mathcal{G}(y_{1:T} | \mathfrak{p}, z, c, \theta_g) = \prod_{t=1}^{T} \mathcal{G}(y_t| y_{1:t-1}, \mathfrak{p}, z, c, \theta_g)
\label{eqn:gen-sample}
\end{equation}




We fine-tune GPT-2 during our pre-training process. We follow the same pre-training procedure as described in Sec.~\ref{sec:spamGANsec}, where we use batches of real sentences from $\mathbb{D}$ and minimize the cross-entropy of the next token conditioned on the preceding ones (see Eqn.~\ref{eqn:genmlegrad}). 
During adversarial training, the generator still acts as a reinforcement learning agent and is trained to maximize the expected rewards (see Eqn.~\ref{eqn:expectreward}). However, in spamGAN-GPT2, we do not use greedy-search, which simply selects the token with the highest probability as the next token, i.e., $y_{t} = argmax\hspace{1mm}\mathcal{G}(y| y_{1:t-1}, \mathfrak{p}, z, c, \theta_g)$ as it may produce sub-optimal results when whole sentences are considered. 
We will instead use a combination of teacher-forcing and top-p sampling strategy~\shortcite{top-p-sampling}. In teacher-forcing, the ground truth is fed as input, i.e., instead of using the generated token $y_t$ at every time-step, the actual token $\hat{y}_t$ is used for sampling from $\mathcal{G}(y| \hat{y}_{1:t-1}, \mathfrak{p}, z, c, \theta_g)$. Teacher-forcing prevents accumulation of errors during the auto-generation process. Further, it can exploit the parallel processing abilities of the transformer architecture improving the training time. We also employ the top-p sampling strategy~\shortcite{top-p-sampling}. Instead of sampling from the distribution $y \sim \mathcal{G}(y| \hat{y}_{1:t-1}, \mathfrak{p}, z, c, \theta_g)$, where $y \in \mathtt{Y}$ considers the probability distribution over all possible tokens in the corpus, top-p sampling, samples from the set of tokens whose cumulative probability exceeds a threshold $p$. These tokens form the top-$p$ vocabulary set $\mathtt{Y}^{(p)} \subset \mathtt{Y}$ as shown in Eqn.~\ref{eqn:top-p-sampling-1}. The probability mass is then redistributed among tokens in $\mathtt{Y}^{(p)}$ according to Eqn.~\ref{eqn:top-p-sampling-2}, where $p' = \sum_{y \in \mathtt{Y}^{(p)} } \mathcal{G}(y| \hat{y}_{1:t-1}, \mathfrak{p}, z, c, \theta_g)$. As the set of tokens in $\mathtt{Y}^{(p)}$ are determined based on the threshold $p$, the size of $\mathtt{Y}^{(p)}$ can dynamically change for every time step. Finally, the next token is sampled from the updated distribution $y \sim \mathcal{G'}(y| \hat{y}_{1:t-1}, \mathfrak{p}, z, c, \theta_g)$, which is obtained using Eqn.~\ref{eqn:top-p-sampling-2}.

\begin{equation}
    \begin{aligned}
        \sum_{y \in \mathtt{Y}^{(p)} } \mathcal{G}(y| \hat{y}_{1:t-1}, \mathfrak{p}, z, c, \theta_g) \ge p
    \end{aligned}
    \label{eqn:top-p-sampling-1}
\end{equation}


\begin{equation}
    \begin{aligned}
       \mathcal{G'}(y| \hat{y}_{1:t-1}, \mathfrak{p}, z, c, \theta_g) = 
       \begin{cases}
       \mathcal{G}(y| \hat{y}_{1:t-1}, \mathfrak{p}, z, c, \theta_g)/p' & \text{if } y \in \mathtt{Y}^{(p)}  \\
       0 & \text{otherwise}
       \end{cases}
    \end{aligned}
    \label{eqn:top-p-sampling-2}
\end{equation}



\begin{figure}[t]
\centering
 \includegraphics[height=6.5cm, keepaspectratio]{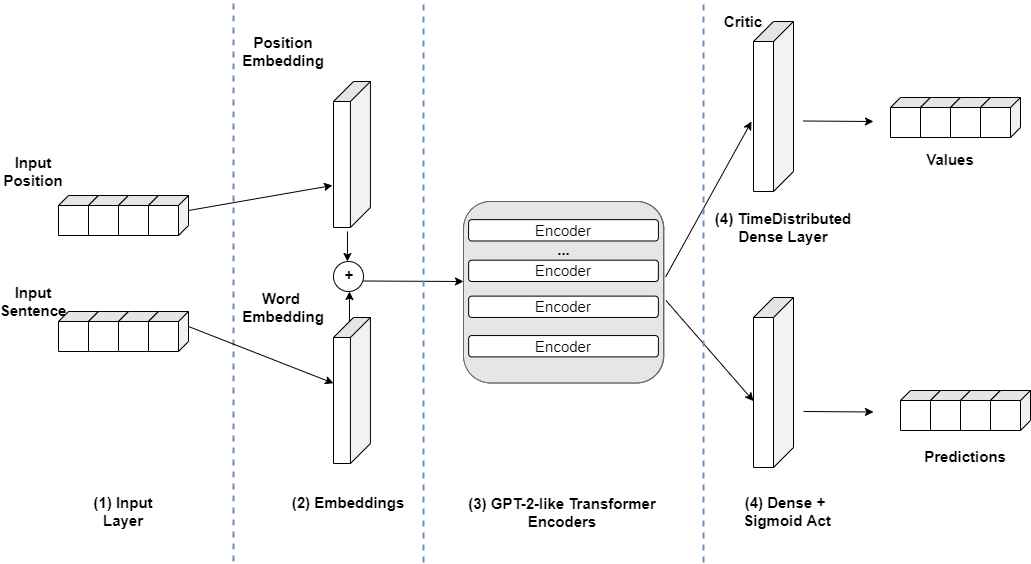}
\caption{spamGAN-GPT2 Discriminator/Classifier Architecture}
\label{fig:spamgangpt2-discrimi}
\vspace{-2mm}
\end{figure}

\vspace{2mm}
\noindent 
{\bf{Discriminator and Classifier.}}
We modify the discriminator architecture in Fig.~\ref{fig:spamgandes} by replacing the original RNN layer with Transformer encoders (see Fig.~\ref{fig:spamgangpt2-discrimi}). Transformer encoders are shown to perform better than decoders for classification tasks. Transformer encoders in our approach have the same architecture as GPT-2 except that they use \textit{un}masked multi-head self-attention layers (the difference can be inferred from Fig.~\ref{fig:gpt2}). Such an architecture allows us to use the same initial pre-trained GPT-2 model for all (generator, discriminator, classifier) components, which is later fine-tuned towards the respective tasks. 

The discriminator $\mathcal{D}$, with parameters $\theta_d$ predicts if a sentence is real (sampled from $P_R$) or fake (produced by the generator). We train the discriminator by minimizing the discriminator loss $\mathcal{L^{(D)}}$(Eqn.~\ref{eqn:dloss}) and discrimination critic loss $\mathcal{L^{(D\text{crit})}}$ (Eqn.~\ref{eqn:dcriticloss}), respectively.
The classifier $\mathcal{C}$ with parameters $\theta_c$ has the same architecture as the discriminator as shown in Fig.~\ref{fig:spamgangpt2-discrimi} and predicts if a sentence belongs to class $c \in \mathbb{C}$. We train classifier by minimizing the classifier loss $\mathcal{L^{C}}$ (Eqn.~\ref{eqn:closs}) and the classification critic loss $\mathcal{L^{(C\text{crit})}}$ (Eqn.~\ref{eqn:ccriticloss}), respectively.

We employ the same pre-training procedure described in Sec.~\ref{sec:spamGANsec} to fine-tune GPT-2 model in all three components. For adversarial training, we follow the same procedure as shown in Alg.~\ref{alg:spamgan}, however, we use the modified generator, discriminator, and classifier architecture and the generator sampling strategy for spamGAN-GPT2 as described above.



\section{Experiments}
\label{sec:experiments}
\subsection{Settings}


We conduct experiments to evaluate the performance of both spamGAN and spamGAN-GPT2 using the TripAdvisor~\shortcite{Ott2011FindingDO} and YelpZip~\shortcite{Rayana2015} datasets. For spamGAN, we perform the following pre-processing steps. All the reviews are converted to lower-case and are tokenized at the word level, with a vocabulary $\mathtt{Y}$ of $10,000$ tokens\footnote{Vocabulary includes all words from labeled data and most frequent words from unlabeled data.}. The maximum sequence length is $T=128$ words. $\mathtt{Y}$ also includes tokens: $\langle\mathtt{start}\rangle$ and $\langle\mathtt{end}\rangle$, which are added to the beginning and at the end of each sentence, respectively; $\langle\mathtt{pad}\rangle$ for padding sentences smaller than $T$ (longer sentences are truncated, ensuring a consistent sentence length); $\langle\mathtt{unk}\rangle$ for replacing out-of-vocabulary words. In spamGAN, the generator consists of $2$ GRU layers of $1,024$ units each and an output dense layer of $10,000$ units with softmax activation. The dimension of the word embedding is $50$. For the generator, the learning rate is $0.001$ and weight decay is $1 \times 10^{-7}$. Gradient clipping is set to a maximum global norm of $5$. The discriminator contains $2$ GRU layers of $512$ units each and a dense layer with a single scalar output along with sigmoid activation. The discrimination critic is implemented as an alternative head using a timedistributed dense layer. For the discriminator, the learning rate is $0.0001$ and weight decay is $1 \times 10^{-4}$. The hyper-parameters for the classifier are the same as that of the discriminator. The generator, discriminator, and classifier are trained using the ADAM optimizer and all use a variational dropout of $0.5$ between the recurrent layers. We set balancing coefficient $\beta=1$ in Eqn.~\ref{eqn:closs}. 

For spamGAN-GPT2, again, all reviews are converted to lower-case. However, the vocabulary size is $50,257$, corresponding to the vocabulary size of GPT-2. We also convert the reviews using a different text representation method called Byte Paring Encoding (BPE) as used in GPT-2. BPE is a middle ground between character and word level language modeling, interpolating between word level inputs for frequent symbol sequences and character level inputs for infrequent symbol sequences~\cite{gpt-2}. For spamGAN-GPT2, the maximum sequence length is increased to $T = 424$, as GPT-2 is capable of processing longer sentences and preserving more semantic information. spamGAN-GPT2 is trained using ADAM optimizer. The learning rate of all three components is $6.25\times10^{-5}$, weight decay for the generator is $1\times10^{-7}$ while it is $1\times10^{-5}$ for discriminator and classifier. The variational dropout used in the generator for the embedding layer, transformer decoder (residual dropout), and output head is $0.2$, $0.1$, and $0.2$, respectively. For the discriminator and classifier, the dropout values used are $0.4$ (embedding layer), $0.1$ (transformer encoder), and $0.4$ (output heads). We use a threshold of $p=0.9$ for the top-p sampling strategy. 



We compare the proposed approaches spamGAN and spamGAN-GPT2 with $2$ supervised methods: 1) DRI-RCNN~\cite{zhang2018dri}, 2) RCNN~\cite{lai2015recurrent} and $2$ semi-supervised methods: 3) Co-Training~\cite{li2011learning} with Naive Bayes, 4) PU Learning~\cite{hernandez2013using} with Naive Bayes (SVM performed poorly). We conduct a wide variety of experiments to show the influence of labeled data. We measure the performance of the approaches when different proportions, i.e., $10\%, 30\%, 50\%, 70\%, 90\%$, and $100\%$ of labeled data are used during training. To analyze the impact of unlabeled data, we also adjoin varying amounts of unlabeled data to the existing labeled data. For example, spamGAN-0 (represents no unlabeled data is used in addition to labeled data during training), spamGAN-50 (50\% unlabeled data is used additionally), spamGAN-70 (70\% unlabeled is used additionally), and spamGAN-100 (100\% unlabeled data additionally). Similarly, we have spamGAN-GPT2-0, spamGAN-GPT2-50, spamGAN-GPT2-70, and spamGAN-GPT2-100.  We also show the performance of our Base RNN classifier (from spamGAN) and Base GPT-2 classifier (from spamGAN-GPT2), which are trained only using the classifier component on real labeled data by minimizing $\mathcal{L^{(C_{\text{R}})}}$ (Eqn.~\ref{eqn:closs}), without using the generator and discriminator. For Co-Train and PU Learn, which are semi-supervised approaches, we show the results when $50\%$ unlabeled data is additionally used during training.
 
 
We measure the accuracy of classification (\% of reviews correctly classified as spam or non-spam) as well as the F1 score (harmonic mean of precision and recall). We also measure the perplexity~\cite{chen1996building} of the generator in generating good quality sentences. The lower the value of perplexity, the better the quality of the generated sentences.

\begin{table*}[h]
\begin{center}
\resizebox{\textwidth}{!}{
\begin{tabular}{l*{6}{l}}
	\toprule
	Method  &10\% Labeled &30\%&50\%&70\%&90\%&100\%\\
	\midrule 
	spamGAN-GPT2-0   &  0.741 $\pm$ 0.02 & {\bf 0.849} $\pm$ 0.02 & {\bf 0.851} $\pm$ 0.01 & {\bf 0.858} $\pm$ 0.02 &	{\bf 0.869} $\pm$ 0.01 & {\bf 0.875} $\pm$ 0.01  \\
	spamGAN-GPT2-50   & {\bf 0.771} $\pm$ 0.04 &   0.811 $\pm$ 0.02 & 0.836 $\pm$ 0.01 & 0.853 $\pm$ 0.01 & 0.862 $\pm$ 0.01 & 0.870 $\pm$ 0.01 \\
	spamGAN-GPT2-70   & 0.751 $\pm$ 0.01 &   0.824 $\pm$ 0.01 & 0.844 $\pm$ 0.02 & 0.848 $\pm$ 0.01 &	0.854 $\pm$ 0.01 & 0.864 $\pm$ 0.01 \\
	spamGAN-GPT2-100   & 0.759 $\pm$ 0.02 &   0.817 $\pm$ 0.02 & 0.831 $\pm$ 0.01 & 0.845 $\pm$ 0.01 & 0.847 $\pm$ 0.01 & 0.861 $\pm$ 0.01  \\
	Base GPT-2 classifier  & 0.734 $\pm$ 0.03 & 0.812 $\pm$ 0.02 & 0.825 $\pm$ 0.01 & 0.841 $\pm$ 0.02 & 0.849 $\pm$ 0.02 & 0.851 $\pm$ 0.02 \\
	spamGAN-0   & 0.700 $\pm$ 0.02 & 0.811 $\pm$ 0.02 & 0.838 $\pm$ 0.01 &	0.845 $\pm$ 0.01 &	0.852 $\pm$ 0.02 &	0.862 $\pm$ 0.01  \\
	spamGAN-50  & 0.678 $\pm$ 0.03 &	0.797 $\pm$ 0.03 &	0.839 $\pm$ 0.02 &	0.845 $\pm$ 0.02 &	0.857 $\pm$ 0.02 &	0.856 $\pm$ 0.01 \\
	spamGAN-70  & 0.695 $\pm$ 0.05 &	0.780 $\pm$ 0.03 &	0.828 $\pm$ 0.02 &	0.850 $\pm$ 0.01 &	0.841 $\pm$ 0.02 &	0.844 $\pm$ 0.02 \\
	spamGAN-100  & 0.681 $\pm$ 0.02 &	0.783 $\pm$ 0.02 &	0.831 $\pm$ 0.01	&0.837 $\pm$ 0.01	& 0.843 $\pm$ 0.02 &	0.845 $\pm$ 0.01\\
	Base RNN classifier  & 0.722 $\pm$ 0.03 &	0.786 $\pm$ 0.02 &	0.791 $\pm$ 0.02 &	0.829 $\pm$ 0.01 &	0.824 $\pm$ 0.02 &	0.827 $\pm$ 0.02\\
	DRI-RCNN & 0.647 $\pm$ 0.10 & 0.757 $\pm$ 0.01 & 0.796 $\pm$ 0.01 & 0.834 $\pm$ 0.18 & 0.835 $\pm$ 0.02 & 0.846 $\pm$ 0.01 \\
	RCNN  & 0.538 $\pm$ 0.09 & 0.665 $\pm$ 0.14 & 0.733 $\pm$ 0.09 & 0.811 $\pm$ 0.03 & 0.834 $\pm$ 0.02 & 0.825 $\pm$ 0.02 \\
	Co-Train (Naive Bayes) & 0.655 $\pm$ 0.01  & 0.740 $\pm$ 0.01 & 0.738 $\pm$ 0.02 & 0.743 $\pm$ 0.01 & 0.754 $\pm$ 0.01 & 0.774 $\pm$ 0.01 \\
	PU Learn (Naive Bayes) & 0.508 $\pm$ 0.02 & 0.713 $\pm$ 0.03 & 0.816 $\pm$ 0.01 & 0.826 $\pm$ 0.01 & 0.838 $\pm$ 0.02 & 0.843 $\pm$ 0.02 \\
	\bottomrule
\end{tabular}}
\end{center}
\caption{TripAdvisor: Accuracy (Mean $\pm$ Std) for Different \% Labeled Data}
\label{tab:trip_accuracy}
\end{table*}

\vspace{-2mm}
\subsection{TripAdvisor}
The TripAdvisor labeled dataset~\cite{Ott2011FindingDO}\footnote{http://myleott.com/op-spam.html} consists of $800$ truthful reviews on Chicago hotels from TripAdvisor and $800$ deceptive reviews obtained from Amazon MechanicalTurk. We remove a small number of duplicate truthful reviews to get a balanced labeled dataset of 1596 reviews. For training, we augment the labeled set with $32,297$ unlabeled TripAdvisor reviews for Chicago hotels\footnote{http://times.cs.uiuc.edu/$\sim$wang296/Data/}~\shortcite{wang2011latent}. The labeled and unlabeled reviews may not follow the same distribution as the labeled non-spam reviews are hand-crafted through MechanicalTurk, while the others are actual reviews from TripAdvisor. We use a $80$-$20$ train-test split on labeled data to perform the evaluation. Experiments are repeated $10$ times and the mean and standard deviation for both accuracy and F1-score are reported. We also show the perplexity of the generator in generating fake sentences. 

\subsubsection{Influence of Labeled Data}

Table.~\ref{tab:trip_accuracy} shows the classification accuracy of all the models on the test data. Our spamGAN-GPT2\footnote{spamGAN-GPT2 refers to all spamGAN-GPT2-0, spamGAN-GPT2-50, spamGAN-GPT2-70, and spamGAN-GPT2-100 models.} model achieves the best performance outperforming spamGAN\footnote{spamGAN refers to all spamGAN-0, spamGAN-50, spamGAN-70, and spamGAN-100 models.} and the other approaches. When merely $10\%$ of the labeled data is used during training, spamGAN-GPT2-0, spamGAN-GPT2-50, spamGAN-GPT2-70 and spamGAN-GPT2-100 achieve an accuracy of $0.741$, $0.771$, $0.751$, and $0.759$, respectively. The base GPT-2 classifier has a lower accuracy ($0.734$) than the spamGAN-GPT2 models, demonstrating the significance of adversarial training. As spamGAN-GPT2-0 is a transformer-based model, it learns much more information than spamGAN-0 (which uses RNNs), outperforming it by $5.86\%$ when $10\%$ labeled data ($128$ reviews) is used. For $10\%$ labeled data, spamGAN-0, spamGAN-50, spamGAN-70, and spamGAN-100 achieve an accuracy of $0.70, 0.678, 0.695$, and $0.681$, respectively. For $10\%$ labeled data, both spamGAN-GPT2-0 and spamGAN-0 outperform other supervised approaches, DRI-RCNN ($0.647$) and R-CNN ($0.538$), as well as semi-supervised approaches Co-train ($0.655$) and PU Learning ($0.508$). Specifically, spamGAN-GPT2-0 outperforms the best performing related work, i.e., Co-train by $13.13\%$. SpamGAN-GPT2-0 and spamGAN-0 also demonstrate better accuracy than the other models when $30\%, 50\%, 70\%, 90\%$, and $100\%$ of labeled data are used during training. 

\begin{figure}[t]
\centering
 \includegraphics[height=6.0cm, keepaspectratio]{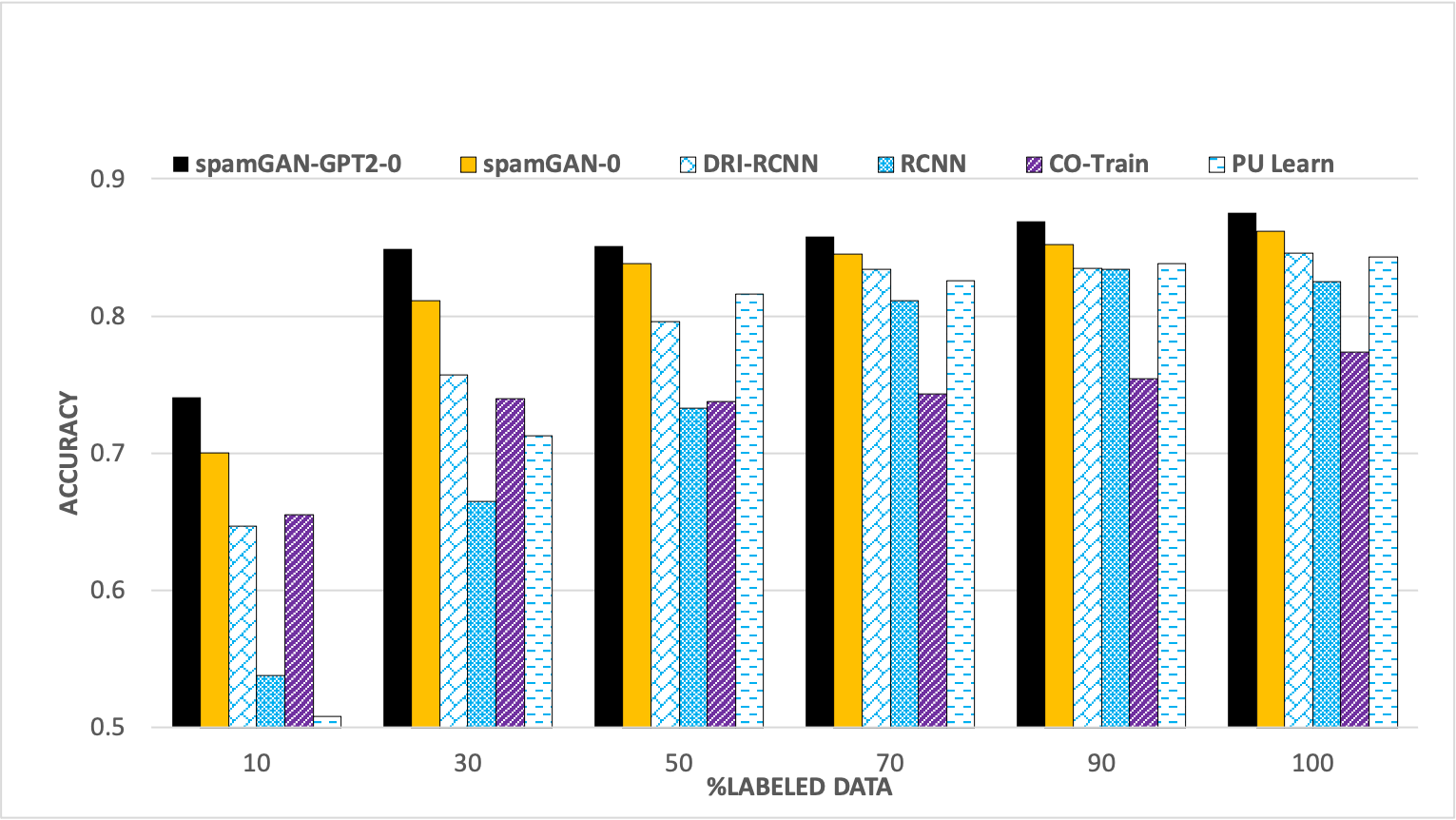}
\caption{Influence of Labeled Data on TripAdvisor Dataset}
 \label{fig:accuracy}
\vspace{-2mm}
\end{figure}

The accuracy of all approaches increases with the increase in the percentage of labeled data. Among all the approaches, spamGAN-GPT2-0 shows the best performance when all proportions of labeled data are considered. However, only for $10\%$ labeled data spamGAN-GPT2-50 performs slightly better than spamGAN-GPT2-0, which can be attributed to the limited availability of labeled data. However, even without unlabeled data, for most proportions of labeled data, spamGAN-GPT2-0 gets good results because the mutual bootstrapping between the generator and classifier allows the classifier to explore beyond the small labeled training set using the fake sentences produced by the generator. We select spamGAN-GPT2-0 and spamGAN-0 as representatives for comparison in Fig.~\ref{fig:accuracy}. The difference in accuracy between spamGAN-GPT2-0 and other approaches reduces as the \% of labeled data increases, obtaining an accuracy of $0.875$ accuracy with $100\%$ labeled data.


\begin{table*}[h]
\begin{center}
\resizebox{\textwidth}{!}{
\begin{tabular}{l*{6}{l}}
	\toprule
	Method &10\% Labeled &30\%&50\%&70\%&90\%&100\%\\
	\midrule 
	spamGAN-GPT2-0  & 0.782 $\pm$ 0.01 & {\bf 0.884} $\pm$ 0.02 & {\bf 0.885} $\pm$ 0.02 & {\bf 0.896} $\pm$ 0.01  & {\bf 0.896} $\pm$ 0.01 & {\bf 0.902} $\pm$ 0.01 \\
	spamGAN-GPT2-50 & {\bf 0.794} $\pm$ 0.04 & 0.857 $\pm$ 0.01 & 0.867 $\pm$ 0.01 & 0.892 $\pm$ 0.01 & 0.895 $\pm$ 0.01 & 0.901 $\pm$ 0.01 \\
	spamGAN-GPT2-70   & 0.784 $\pm$ 0.02 & 0.863 $\pm$ 0.02 & 0.882 $\pm$ 0.02 & 0.883 $\pm$ 0.01 & 0.887 $\pm$ 0.01 & 0.899 $\pm$ 0.01 \\
	spamGAN-GPT2-100 & 0.786 $\pm$ 0.03  & 0.855 $\pm$ 0.02 & 0.872 $\pm$ 0.01 & 0.881 $\pm$ 0.01 & 0.880 $\pm$ 0.02 & 0.893 $\pm$ 0.01 \\
	Base GPT-2 classifier  & 0.754 $\pm$ 0.04 &	0.850 $\pm$ 0.02 & 0.865 $\pm$ 0.01 & 0.875 $\pm$ 0.02 & 0.888 $\pm$ 0.01 & 0.889 $\pm$ 0.02 \\
	spamGAN-0   & 0.718 $\pm$ 0.02	& 0.812 $\pm$ 0.02 &	0.840 $\pm$ 0.01 &	0.848 $\pm$ 0.02 &	 0.854 $\pm$ 0.02 &	0.868 $\pm$ 0.01  \\
	spamGAN-50  & 0.674 $\pm$ 0.05	&0.797 $\pm$ 0.03 &0.843 $\pm$ 0.01	 &0.848 $\pm$ 0.02	& 0.860 $\pm$ 0.02 &	0.863 $\pm$ 0.01 \\
	spamGAN-70  & 0.702 $\pm$ 0.05	&0.784 $\pm$ 0.03	&0.830 $\pm$ 0.02	&0.856 $\pm$ 0.01	&0.848 $\pm$ 0.02 &	0.854 $\pm$ 0.01 \\
	spamGAN-100  & 0.684 $\pm$ 0.03 &	0.788 $\pm$ 0.03 &	0.839 $\pm$ 0.02 &	0.844 $\pm$ 0.01	&0.846 $\pm$ 0.02 &	0.850 $\pm$ 0.01 \\
	Base RNN classifier  & 0.731 $\pm$ 0.03	&0.795 $\pm$ 0.03	&0.803 $\pm$ 0.02	&0.829 $\pm$ 0.01	&0.832 $\pm$ 0.02	&0.838 $\pm$ 0.02 \\
	DRI-RCNN  & 0.632 $\pm$ 0.07 & 0.754 $\pm$ 0.02 & 0.779 $\pm$ 0.00 & 0.812 $\pm$ 0.03 & 0.817 $\pm$ 0.03 & 0.833 $\pm$ 0.02\\
	RCNN   & 0.638 $\pm$ 0.01 & 0.715 $\pm$ 0.01 & 0.754 $\pm$ 0.02 & 0.776 $\pm$ 0.05 & 0.820 $\pm$ 0.03 & 0.833 $\pm$ 0.02\\
	Co-Train (Naive Bayes) & 0.637 $\pm$ 0.02 & 0.698 $\pm$ 0.01 &  0.680 $\pm$ 0.02 & 0.677 $\pm$ 0.01& 0.712 $\pm$ 0.01 & 0.726 $\pm$ 0.01 \\
	PU Learn (Naive Bayes) & 0.050 $\pm$ 0.02 & 0.636 $\pm$ 0.05 & 0.815 $\pm$ 0.02 & 0.837 $\pm$ 0.02 & 0.844 $\pm$ 0.02 & 0.852 $\pm$ 0.01 \\
	\bottomrule
\end{tabular}}
\end{center}
\vspace{-2.5mm}
\caption{TripAdvisor: F1-Score (Mean $\pm$ Std) for Different \% Labeled Data}
\vspace{-2mm}
\label{tab:f1}
\end{table*}

Table.~\ref{tab:f1} shows the F1-score. Again, in general, spamGAN-GPT2-0 achieves the best performance when compared to all other approaches. Specifically, when the percentage of the labeled data is small ($10$\%), spamGAN-GPT2-0 achieves an F1-score of $0.782$ while spamGAN-0 achieves $0.718$. When the proportion of the labeled data increases, the F1-score also increases for all the models, with spamGAN-GPT2-0 and spamGAN-0 achieving $0.902$ and $0.868$, respectively when $100$\% labeled data are considered. The higher F1 score for spamGAN-GPT2 is also attributed to the improved recall when compared to spamGAN. It also outperforms other approaches DRI-RCNN, R-CNN, Co-train, and PU Learning.



\subsubsection{Influence of Unlabeled Data}
\begin{figure}[h]
\vspace{-2mm}
\centering
 \includegraphics[height=6.0cm, keepaspectratio]{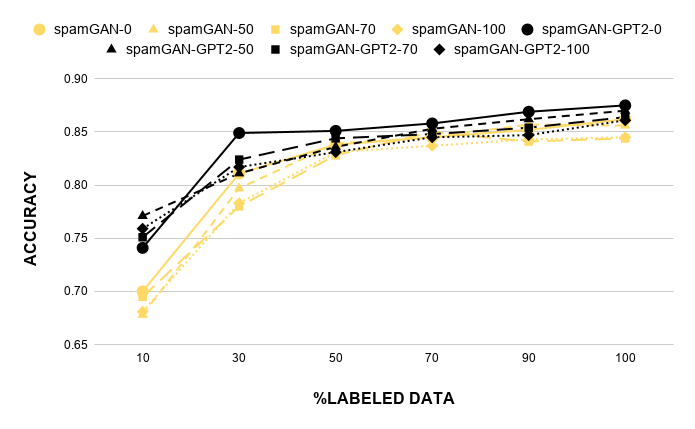}
\caption{Influence of UnLabeled Data on TripAdvisor Dataset}
 \label{fig:trip_f1}
\vspace{-2mm}
\end{figure}

While unlabeled data is used to augment the classifier's performance, Fig.~\ref{fig:trip_f1} shows that both for spamGAN and spamGAN-GPT2, accuracy slightly decreases when the percentage of unlabeled data increases, i.e., accuracy for spamGAN-GPT2-100 is less than spamGAN-GPT2-0 and spamGAN-100 is less than spamGAN-0. For the TripAdvisor dataset, as the unlabeled data with $32,297$ reviews is much larger than the labeled dataset with $1596$ reviews, the generator does not entirely learn the importance of the sentence classes during pre-training (where the unlabeled sentence classes are drawn from a random distribution). This causes problems for the classifier during adversarial training. However, with no unlabeled data, the generator easily learns to generate sentences conditioned on classes, paving way for the mutual bootstrapping between classifier and generator. We also attribute the drop in performance to the difference in distribution of data between unlabeled TripAdvisor reviews and handcrafted spam reviews from MechanicalTurk as the influence of unlabeled data is dependant on assumptions about data distribution~\shortcite{2008statistical}. 


\subsubsection{Perplexity of Generated Sentence}
\begin{figure}
\centering
\subfloat[spamGAN-GPT2]{
  \includegraphics[height=4.7cm, keepaspectratio]{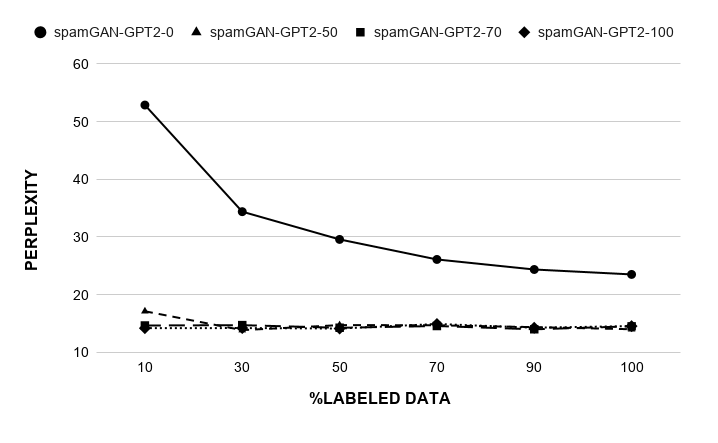}
  \label{fig:trip_gpt2_perplexity}}
\subfloat[spamGAN]{
  \includegraphics[height=4.7cm, keepaspectratio]{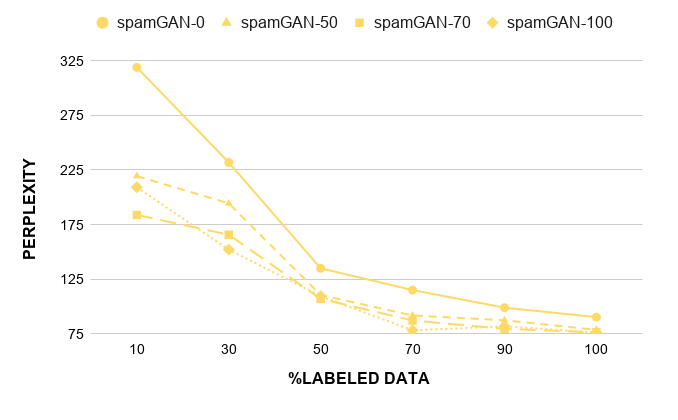}
    \label{fig:trip_spam_perplexity}}
\caption{Perplexity of Generator on TripAdvisor Dataset}
\label{fig:trip_perplexity}
\end{figure}

We also compute the perplexity of sentences produced by the generator (the lower the value the better). Fig.~\ref{fig:trip_perplexity} shows that as the percentage of unlabeled data increases, the perplexity of the sentences decreases both for spamGAN-GPT2 and spamGAN. Though unlabeled data do not improve the classification results (as shown in Fig.~\ref{fig:trip_f1}), they improve the quality of the synthetic sentences generated by the generator. In Fig.~\ref{fig:trip_gpt2_perplexity}, when $10\%$ labeled data is used, spamGAN-GPT2-0 obtains a perplexity of $52.8$ and spamGAN-GPT2-100 obtains a better perplexity of $14.2$. In Fig.~\ref{fig:trip_spam_perplexity}, spamGAN-0 and spamGAN-100 obtain a perplexity of $318.8$ and $209.2$, respectively. Thus, with $10\%$ labeled data, spamGAN-GPT2-0 is able to improve the quality of sentences generated by $83.4$\% when compared to spamGAN-0. GPT-2 is well known for its capability to handle natural language generation tasks. Compared to the traditional RNN architecture in spamGAN, the use of transformer-based GPT-2 architecture in spamGAN-GPT2 allows to capture better semantic information, thereby significantly improving the quality of the sentences generated by the generator. 

We also note that there is a large improvement in perplexity between spamGAN-GPT2-0 and spamGAN-GPT2-50, due to the unlabeled data additionally considered during training. However, when more unlabeled data (spamGAN-GPT2-70/90/100) is added, the perplexity still lingers around $15$. This is because using $50\%$ unlabeled data, i.e., $16148$ unlabeled reviews already provides sufficient semantic information for GPT-2 to capture.

Fig.~\ref{fig:trip_f1} and Fig.~\ref{fig:trip_perplexity} show that using unlabeled data improves the generator in producing realistic sentences but does not fully help to differentiate between the class labels. This again can be attributed to the difference in data distribution between the labeled and unlabeled data and the difference in the number of labeled and unlabeled reviews considered. 

\subsection{YelpZip}
We also conduct experiments on the Yelp review dataset~\shortcite{Rayana2015} consisting of $608,598$ restaurant reviews out of which $13.22$\% are spam. To balance the dataset, we sample $40,219$ spam and $40,219$ non-spam reviews. We use the remaining $80,438$ reviews as unlabeled samples by removing the ground truth labels. These unlabeled samples have an equal number of spam and non-spam reviews. As the labeled and unlabeled data are derived from the same dataset, both of them follow the same data distribution. We also use an equal number of labeled and unlabeled reviews. We use a 80-20 train-test split on labeled data to perform the evaluation. Experiments are repeated $5$ times for this larger dataset and the mean and standard deviation for accuracy and F1-score are reported. However, for spamGAN-GPT2, we run the experiments only once due to the long training time and the overwhelming cost of renting AWS EC2 instances necessary to accommodate the GPT-2 model (see Sec.~\ref{sec:time} for details). Also, YelpZip is much larger than TripAdvisor: labeled data is $50$ times larger, and unlabeled data is $2.5$ times larger. 

\subsubsection{Influence of Labeled Data}

 \begin{figure}[h]
\centering
 \includegraphics[height=6.0cm, keepaspectratio]{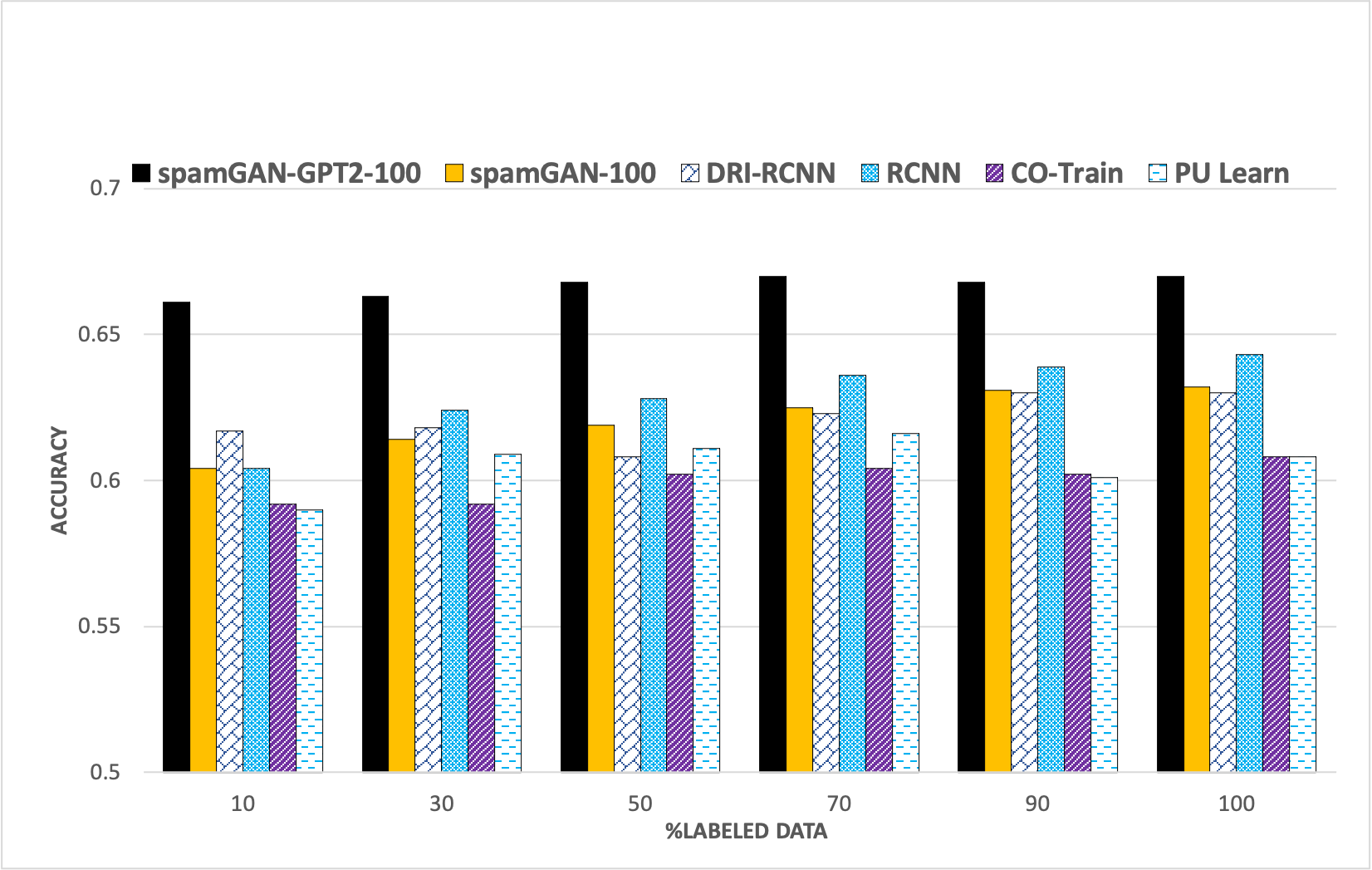}
\caption{Influence of Labeled Data on YelpZip Dataset}
 \label{fig:acc-yelp}
\vspace{-2mm}
\end{figure} 

Table.~\ref{tab:yelp_accuracy} shows the classification accuracy of the YelpZip data. When the labeled data is limited to $10\%$, spamGAN-GPT2-0, spamGAN-GPT2-50, spamGAN-GPT2-70, and spamGAN-GPT2-100 achieve an accuracy of $0.647$, $0.653$, $0.655$, and $0.661$ respectively, which outperform other supervised/semi-supervised approaches as well as the spamGAN model. The performance of spamGAN for YelpZip is moderate with an accuracy of $0.604$ for $10$\% labeled when all unlabeled data are considered. This is only slightly better than RCNN ($0.604$), Co-Train ($0.592$), and PU-Train ($0.590$), though not as good as DRI-RCNN ($0.617$). spamGAN-GPT2-100 outperforms spamGAN-100 by $9.44$\% in terms of accuracy and the best performing state-of-the-art approach DRI-RCNN by $7.13$\%. 

\begin{table*}[t]
\begin{center}
\resizebox{\textwidth}{!}{
\begin{tabular}{l*{6}{l}}
	\toprule
	Method  &10\% Labeled &30\%&50\%&70\%&90\%&100\%\\
	\midrule 
	spamGAN-GPT2-0\%    & 0.647 $\pm$ 0.007 & 0.654 $\pm$ 0.008 & 0.660 $\pm$ 0.000  & 0.662 $\pm$ 0.000 & 0.665 $\pm$ 0.000 & 0.667 $\pm$ 0.000   \\
	spamGAN-GPT2-50\%    &  0.653 $\pm$ 0.002 & 0.657 $\pm$ 0.003 & 0.664 $\pm$ 0.000 & 0.657 $\pm$ 0.000 & 0.666 $\pm$ 0.000 & 0.666 $\pm$ 0.000 \\ 
	spamGAN-GPT2-70\%    &  0.655 $\pm$ 0.000 & 0.660 $\pm$ 0.000 & 0.666 $\pm$ 0.000 & 0.660 $\pm$ 0.000  & 0.665 $\pm$ 0.001 & {\bf 0.673} $\pm$ 0.002 \\
	spamGAN-GPT2-100\%    &  {\bf 0.661} $\pm$ 0.000 & {\bf 0.663} $\pm$ 0.000 & {\bf 0.668} $\pm$ 0.006 & {\bf 0.670} $\pm$ 0.000 & {\bf 0.668} $\pm$ 0.000 & 0.670 $\pm$ 0.003  \\
	Base GPT-2 classifier  &  0.642 $\pm$ 0.009 &	0.654 $\pm$ 0.005 &	0.659 $\pm$ 0.000 &	0.660 $\pm$ 0.000 & 0.661 $\pm$ 0.000 & 0.663 $\pm$ 0.000 \\
	spamGAN-0\%   & 0.610 $\pm$ 0.001 &  0.613 $\pm$ 0.001 & 0.620 $\pm$ 0.001 & 0.624 $\pm$ 0.003 &	 0.631 $\pm$ 0.001  & 0.634 $\pm$ 0.001  \\
	spamGAN-50\%  & 0.605 $\pm$ 0.008 &	0.612 $\pm$ 0.001 &	0.616 $\pm$ 0.001 &	0.624 $\pm$ 0.001  &	0.631 $\pm$ 0.001 & 0.634 $\pm$ 0.002 \\
	spamGAN-70\%  & 0.607 $\pm$ 0.001 &	0.612 $\pm$ 0.001 &	0.618 $\pm$ 0.003 &	0.626 $\pm$ 0.002  &	0.633 $\pm$ 0.001 & 0.634 $\pm$ 0.001 \\
	spamGAN-100\%  & 0.604 $\pm$ 0.001  & 0.614 $\pm$ 0.001 &	0.619 $\pm$ 0.001 &	0.625 $\pm$ 0.001 &	0.631 $\pm$ 0.003 & 0.632 $\pm$ 0.002 \\
	Base RNN classifier  &  0.616 $\pm$ 0.002 &	0.618 $\pm$ 0.002 &	0.619 $\pm$ 0.003 &	 0.622 $\pm$ 0.001 &	0.625 $\pm$ 0.001 &  0.626 $\pm$ 0.002 \\
	DRI-RCNN & 0.617 $\pm$ 0.001 &	0.618 $\pm$ 0.001 &	0.608 $\pm$ 0.001 &	0.623 $\pm$ 0.001 &	0.630 $\pm$ 0.001 & 0.630 $\pm$ 0.001 \\
	RCNN  & 0.604 $\pm$ 0.001 &	0.624 $\pm$ 0.001 &	0.628 $\pm$ 0.001 &	0.636 $\pm$ 0.001 &	0.639 $\pm$ 0.001 & 0.643 $\pm$ 0.001 \\
	Co-Train (Naive Bayes) & 0.592 $\pm$ 0.001  & 0.592 $\pm$ 0.001 & 0.602 $\pm$ 0.001 & 0.604 $\pm$ 0.001 & 0.602 $\pm$ 0.001 & 0.608 $\pm$ 0.001 \\
	PU Learn (Naive Bayes) & 0.590 $\pm$ 0.001  &	0.609 $\pm$ 0.001 &	0.611 $\pm$ 0.001 &	0.616 $\pm$ 0.001 &	0.601 $\pm$ 0.001 & 0.608 $\pm$ 0.001 \\
	\bottomrule
\end{tabular}}
\end{center}
\vspace{-2.5mm}
\caption{YelpZip: Accuracy (Mean $\pm$ Std) for Different \% Labeled Data}
\vspace{-2.5mm}
\label{tab:yelp_accuracy}
\end{table*}


As the percentage of labeled data increases, the accuracy of all the approaches increases. We choose spamGAN-GPT2-100 and spamGAN-100 as representatives for the performance comparison shown in Fig.~\ref{fig:acc-yelp}. spamGAN-GPT2-100 outperforms all other approaches including spamGAN achieving accuracy of $0.670$ when all $100\%$ labeled data are used during training. Though spamGAN achieves a reasonable accuracy, it is not able to always outperform the other supervised approaches like RCNN and DRI-RCNN. The average length of YelpZip reviews is $186$ (it is $132$ for TripAdvisor), which cannot be effectively handled by RNN-based approaches such as spamGAN. The GPT-2 architecture, however, is able to handle longer sentences, allowing spamGAN-GPT2 to achieve good accuracy.

Table.~\ref{tab:yelp_f1} shows the F1-score. The F1-scores increase with the proportion of labeled data used for training. For $10\%$ labeled data, The F1-scores of spamGAN-GPT2-0, spamGAN-GPT2-50, spamGAN-GPT2-70, and spamGAN-GPT2-100 are $0.653$, $0.660$, $0.671$, and $0.682$, respectively. spamGAN-GPT2-100 outperforms all other approaches DRI-RCNN ($0.654$), RCNN ($0.637$), Co-Train ($0.644$), and PU-Train ($0.654$) as well as spamGAN. spamGAN-100 achieves an F1-score of $0.610$ when $10\%$ labeled data is considered.



\begin{table*}[t]
\begin{center}
\resizebox{\textwidth}{!}{
\begin{tabular}{l*{6}{l}}
	\toprule
	Method &10\% Labeled &30\%&50\%&70\%&90\%&100\%\\
	\midrule 
    spamGAN-GPT2-0\%    & 0.653 $\pm$ 0.006 & 0.649 $\pm$ 0.033 & 0.650 $\pm$ 0.000 & 0.686 $\pm$ 0.000 & 0.653 $\pm$ 0.000 & 0.683 $\pm$ 0.000    \\
	spamGAN-GPT2-50\%    &  0.660 $\pm$ 0.011 & 0.669 $\pm$ 0.003 & 0.653 $\pm$ 0.000 & 0.638 $\pm$ 0.000 & 0.649 $\pm$ 0.000 & 0.691 $\pm$ 0.000   \\
	spamGAN-GPT2-70\%    & 0.671  $\pm$ 0.000 & 0.671 $\pm$ 0.000 & 0.672 $\pm$ 0.000 & 0.669 $\pm$ 0.000 & 0.662 $\pm$ 0.007 & {\bf 0.698} $\pm$ 0.005   \\
	spamGAN-GPT2-100\%    & {\bf 0.682} $\pm$ 0.000 & {\bf 0.674} $\pm$ 0.000 & {\bf 0.694} $\pm$ 0.003 & {\bf 0.687} $\pm$ 0.000 & {\bf 0.665} $\pm$ 0.000 & 0.695 $\pm$ 0.009   \\
	Base GPT-2 classifier  & 0.622 $\pm$ 0.003 &	0.655 $\pm$ 0.034 & 0.650 $\pm$ 0.000 & 0.678 $\pm$ 0.000 & 0.644	$\pm$ 0.000 & 0.650 $\pm$ 0.000	 \\
	spamGAN-0\%  &  0.612 $\pm$ 0.001 	& 0.621 $\pm$ 0.003 & 0.627 $\pm$ 0.005 &	0.629 $\pm$ 0.001 &	 0.630 $\pm$ 0.001 &	0.631 $\pm$ 0.002  \\
	spamGAN-50\%  & 0.611 $\pm$ 0.001 & 0.612 $\pm$ 0.004 & 0.620 $\pm$ 0.003 &	0.624 $\pm$ 0.001  &	 0.620 $\pm$ 0.001 &	0.630 $\pm$ 0.001  \\
	spamGAN-70\%  & 0.607 $\pm$ 0.001 & 0.622 $\pm$ 0.002 &	0.627 $\pm$ 0.002 &	0.625 $\pm$ 0.003 &	 0.618 $\pm$ 0.001 &	0.631 $\pm$ 0.001  \\
	spamGAN-100\%  & 0.610 $\pm$ 0.002	& 0.624 $\pm$ 0.001 &	0.626 $\pm$ 0.001 &	0.625 $\pm$ 0.001 &	 0.625 $\pm$ 0.001 &	0.631 $\pm$ 0.001 \\
	Base RNN classifier  & 0.620 $\pm$ 0.007	& 0.622 $\pm$ 0.007 &	 0.631 $\pm$ 0.014 &	0.631 $\pm$ 0.006 &	0.630 $\pm$ 0.001 & 0.630 $\pm$ 0.011  \\
	DRI-RCNN  & 0.654 $\pm$ 0.013	& 0.654 $\pm$ 0.001 &	0.623 $\pm$ 0.027 &	0.634 $\pm$ 0.001 &	0.650  $\pm$ 0.003 &	0.652 $\pm$ 0.001  \\
	RCNN   & 0.637 $\pm$ 0.050 & 0.646 $\pm$ 0.001 & 0.650 $\pm$ 0.001 &	0.658 $\pm$ 0.001 &	0.666 $\pm$ 0.021 & 0.659 $\pm$ 0.014  \\
	Co-Train (Naive Bayes) & 0.644 $\pm$ 0.001 & 0.658 $\pm$ 0.001 &  0.655 $\pm$ 0.001 & 0.659 $\pm$ 0.001 & 0.659 $\pm$ 0.001 & 0.665 $\pm$ 0.001 \\
	PU Learn (Naive Bayes) & 0.654 $\pm$ 0.001	& 0.656 $\pm$ 0.001 &	0.552 $\pm$ 0.001 &	0.584 $\pm$ 0.001 &	0.527  $\pm$ 0.001 &	0.539 $\pm$ 0.001  \\
	\bottomrule
\end{tabular}}
\end{center}
\vspace{-2.5mm}
\caption{YelpZip: F1-Score (Mean $\pm$ Std) for Different \% Labeled Data}
\vspace{-1.5mm}
\label{tab:yelp_f1}
\end{table*}


  \begin{figure}[h]
\centering
 \includegraphics[height=6.0cm, keepaspectratio]{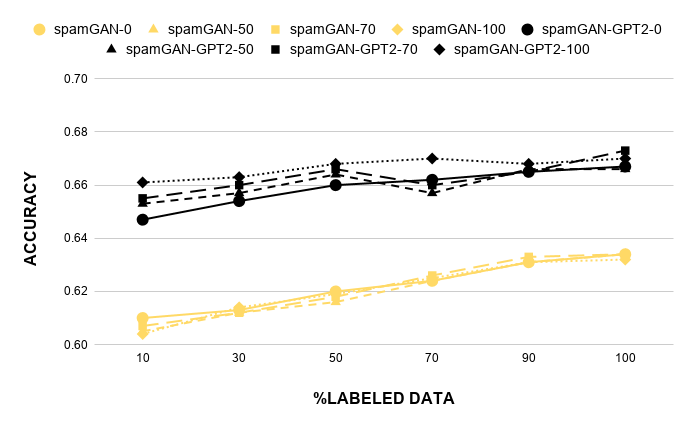}
\caption{Influence of UnLabeled Data on YelpZip Dataset}
\label{fig:yelp_acc}
\vspace{-2mm}
\end{figure}

\subsubsection{Influence of Unlabeled Data}

Unlike TripAdvisor, Fig.~\ref{fig:yelp_acc} shows that the spamGAN-GPT2 achieves better accuracy as the percentage of unlabeled data increases, with spamGAN-GPT2-100 achieving the overall best performance for the YelpZip dataset. This shows us that when labeled and unlabeled data follow the same distribution and are fairly equal in number, the unlabeled data help the generator of spamGAN-GPT2 learn the importance of the assigned sentence classes during training. Thereby, the generator is able to generate better spam-distinguishable sentences conditioned on classes with more unlabeled data. SpamGAN shows a slight improvement in accuracy with more unlabeled data, however, the improvement is not significant.



\subsubsection{Perplexity of Generated Sentences}

\begin{figure}[h]
\centering
\hspace{-2mm}
 \subfloat[spamGAN-GPT2]{
  \includegraphics[height=4.7cm, keepaspectratio]{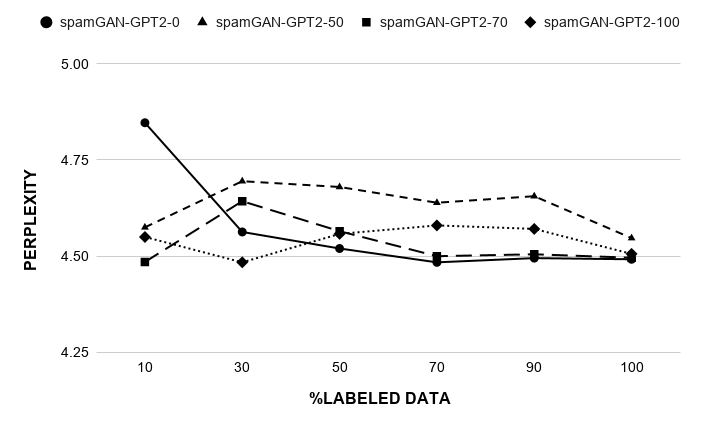}
   \label{fig:yelp_gpt2_perplexity}
}
\subfloat[spamGAN]{
  \includegraphics[height=4.7cm, keepaspectratio]{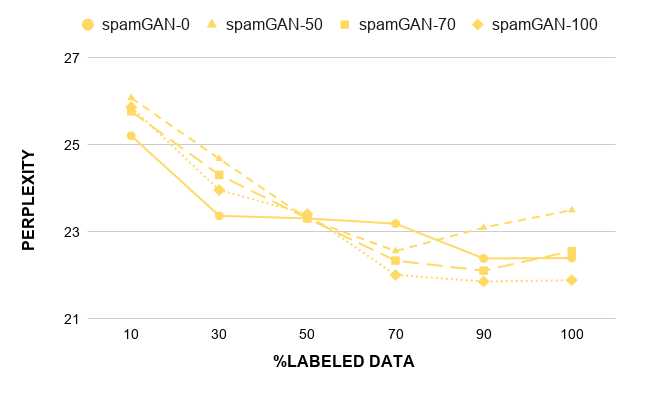}
   \label{fig:yelp_spam_perplexity}
}
\caption{Perplexity of Generator on YelpZip Dataset}
\label{fig:yelp_perplexity}
\end{figure}

Fig.~\ref{fig:yelp_perplexity} shows perplexity of the YelpZip sentences generated by the generator of both spamGAN-GPT2 and spamGAN. In Fig.~\ref{fig:yelp_gpt2_perplexity}, spamGAN-GPT2-100 achieves a perplexity of $4.506$ when all $100$\% labeled data are used during training. Perplexity values fluctuate between $4.5 \sim 4.8$ for different proportions of unlabeled data. In Fig.~\ref{fig:yelp_spam_perplexity}, spamGAN-100 achieves a perplexity of $21.89$. These perplexity values are much lower than those for the TripAdvisor dataset. This can be attributed to the consistency in data distribution between labeled and unlabeled reviews in YelpZip dataset. Furthermore, YelpZip dataset is much larger than TripAdvisor. With more reviews, more semantic information can be obtained by the generator, resulting in better quality sentences. In YelpZip, spamGAN-GPT2-100 improves perplexity of the sentences generated when compared to spamGAN-100 by $82.46$\%.

\subsection{Time Complexity} \label{sec:time}
For experiments on TripAdvisor, we used GeForce RTX 2070 GPU. The training time of spamGAN-GPT2 was $9$ hours for one model for a single run when all labeled and unlabeled data were used, whereas for spamGAN it took only $3$ hours. Although spamGAN-GPT2 requires more computational time when compared to spamGAN, the performance gains are higher in terms of accuracy and perplexity as described in previous sub-sections. For the larger YelpZip dataset, we used g4dn.2xlarge AWS EC2 instance with Tesla T4 to train spamGAN-GPT2, which took $200$ hours per model for a single train run.

\section{Conclusion and Future Work}\label{sec:conclusion}

We propose spamGAN-GPT2, an approach for detecting opinion spam with limited labeled data. spamGAN-GPT2 consists of $3$ components, the generator, discriminator, and classifier, which are adversarially trained to eventually classify reviews as spam or non-spam. A reinforcement learning framework is also employed to address the sparse rewards problem that arises during sentence generation. spamGAN-GPT2 leverages the pre-trained transformer-based GPT-2 model and achieves better classification results than the RNN based spamGAN approach. spamGAN-GPT2 also uses the top-p sampling strategy along with teacher-forcing to avoid sub-optimal sentences during the auto-regressive sentence generation process. Experiments on TripAdvisor and YelpZip show that spamGAN-GPT2 outperforms the state-of-the-art supervised and semi-supervised spam detection techniques, especially when labeled data is limited. Results also show that spamGAN-GPT2 can generate better quality synthetic sentences that can be used to augment the training data.

For future work, we plan to improve the performance of our model by updating the conditional sentence generation process. Instead of drawing class labels from a random distribution, we plan to label the unlabeled reviews using the pre-trained classifier. This will avoid incorrectly labeling the unlabeled reviews, which can further improve the classification accuracy. As the proposed model can be applied to any text classification task, we plan to conduct experiments on other applications such as sentiment analysis.


\acks{This work has been partially supported by the AWS Cloud Credits for Research program.}

\bibliographystyle{theapa}

\end{document}